\documentclass[journal]{IEEEtran}
\usepackage{cite}

\ifCLASSINFOpdf
  \usepackage[pdftex]{graphicx}
\else
\fi
\usepackage{algorithmic}
\ifCLASSOPTIONcompsoc
  \usepackage[caption=false,font=normalsize,labelfont=sf,textfont=sf]{subfig}
\else
  \usepackage[caption=false,font=footnotesize]{subfig}
\fi

\usepackage{amssymb}
\usepackage{amsmath}
\usepackage{autobreak}
\usepackage{amsfonts}

\usepackage{url}
\usepackage{multirow}
\usepackage{bm}

\usepackage{color, colortbl}
\usepackage{xcolor}
\definecolor{lightgray}{gray}{0.1}
\definecolor{orange}{rgb}{1,0.5,0}

\usepackage{fancyhdr}
\begin{document}
%

\title{Uncovering Interpretable Internal States of Merging Tasks at Highway On-Ramps for Autonomous Driving Decision-Making}

\author{Huanjie~Wang,
        Wenshuo~Wang,~\IEEEmembership{Member,~IEEE,}
        Shihua~Yuan,
        and~Xueyuan~Li
\thanks{This work is supported by the National Natural Science Foundation of China (Grant No. U1864210). (\textit{Corresponding Authors: Wenshuo Wang and Shihua Yuan})}

\thanks{H. Wang is with the School of Mechanical Engineering, Beijing Institute of Technology, Beijing 10081, China (e-mail: wanghj\_815@163.com).

W. Wang is with the University of California at Berkeley, Berkeley, CA 94720 USA (e-mail: wwsbit@gmail.com).}

\thanks{S. Yuan and X. Li are with the School of Mechanical Engineering, Beijing Institute of Technology, Beijing 10081, China (e-mail: yuanshihua@bit.edu.cn).}
}


%

\maketitle

\thispagestyle{fancy}

\begin{abstract}

Humans make daily routine decisions based on their internal states in intricate interaction scenarios. This paper presents a probabilistically reconstructive learning approach to identify the internal states of multi-vehicle sequential interactions when merging at highway on-ramps. We treated the merging task's sequential decision as a dynamic, stochastic process and then integrated the internal states into an HMM-GMR model, a probabilistic combination of an extended Gaussian mixture regression (GMR) and hidden Markov models (HMM). We also developed a variant expectation-maximum (EM) algorithm to estimate the model parameters and verified it based on a real-world data set. Experiment results reveal that three interpretable internal states can semantically describe the interactive merge procedure at highway on-ramps. This finding provides a basis to develop an efficient model-based decision-making algorithm for autonomous vehicles (AVs) in a partially observable environment.

\textit{Note to Practitioners}---Model-based learning approaches have obtained increasing attention in decision-making design due to their stability and interpretability. This paper was built upon the two facts: (1) Intelligent agents can only receive partially observable environment information directly through their equipped sensors in the real world; (2) Humans mainly utilize the internal states and associated dynamics inferred from observations to make proper decisions in complex environments. Similarly, AVs need to understand, infer, anticipate, and exploit the internal states of dynamic environments. Applying probabilistic decision-making models to AVs requires updating the internal states' beliefs and associated dynamics after getting new observations. The designed and verified emission model in HMM-GMR provides a modifiable functional module for online updates of the associated internal states. Experiment results based on the real-world driving dataset demonstrates that the internal states extracted using HMM-GMR can represent the dynamic decision-making process semantically and make an accurate prediction. 

\end{abstract}


%
\IEEEpeerreviewmaketitle

\section{Introduction}
%
%
%
%

\IEEEPARstart{T}{aking} an efficient and safe merge at highway on-ramps is a daily-routine but challenging task for humans and autonomous agents in the real world \cite{lin2019decision}. Near 30,000 highway merging collisions occurred per year in the USA\cite{national2018summary}. Typical highway traffic issues such as oscillations, congestion, and speed breakdown are arising incrementally due to inefficient collaborations between the ego vehicle and its surroundings \cite{marczak2013merging}. Thus, taking insights into humans' cooperative merging processes in a changing context becomes indispensable to make a safer, more efficient decision for autonomous vehicles (AVs). 

\begin{figure}[t]
\centering
\includegraphics[width=\linewidth]{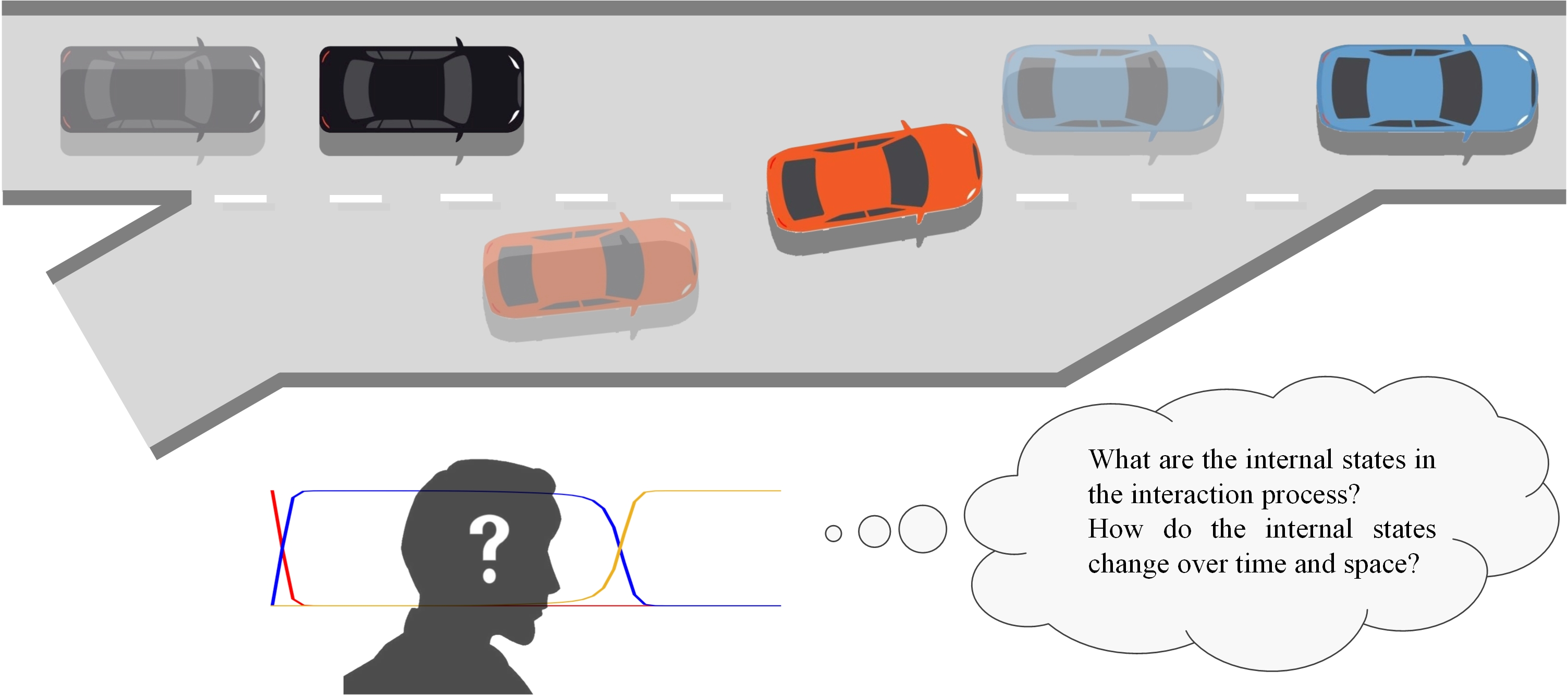}
\caption{The interaction of the ego vehicle (red) with its nearby surroundings (black and blue) when merging into the highway from on-ramps. The ego vehicle makes a proper decision based on its \textit{internal} model of the dynamic environment.}
\label{fig:Internal_state}
\end{figure}

Humans can interact with non-stationary, separately controlled, and partially observable agents seamlessly but do not need to explicitly model each other's strategy during the implementation of complex interaction processes \cite{rubinstein1998modeling}. Inspired by this, artificial agents (such as AVs) should make high-level strategy representations based on their observation of other agents' low-level actions. The high-level strategy representation is implicit, known as \textit{internal states}, which are usually changing over time \cite{xie2020learning}. The agents then take actions based on their previous choices of different plans or strategies. The low-level action is measurable; however, the other agents' planning and internal state changes are unobservable but significant for the sequential decision process. When merging at highway on-ramps (as shown in Fig. \ref{fig:Internal_state}), the human agent can directly detect the contextual traffic changes through their sensory dynamics, but not for the underlying states (such as intents) of the surrounding vehicles, which requires inference from the directly perceived signals. With this situation, the decision to merge in congested traffic involves a tremendous amount of cooperation, courtesy, and risk-taking and must explicitly consider the internal states' change and influences \cite{choudhury2010dynamic}.

The introduction of internal states allows to mathematically formulate many existing decision-making problems via solvable, tractable mathematical models. A typical, popular decision-making model is built upon the Markov decision process (MDP), which basically describes the sequential decision-making mechanisms in complex interactive environments such as the merging procedure at highway on-ramps. One of its derivations, called partially observable MDP (POMDP), has been widely used to formulate the decision-making problem whose partial states are unobservable. Research in \cite{hubmann2018belief} defined the high-level discrete (internal) states of interactive merge behavior to formulate the decision-making into a solvable POMDP problem. Another typical model built upon MDP is deep reinforcement learning (Deep RL), which increases attention in the decision-making of autonomous driving when combined with deep neural networks \cite{forbes2002reinforcement, sallab2017deep, zhang2018coarse, wang2021tactical}. However, the Deep RL strategy usually lacks interpretability and generalizability and can only adapt to well-defined and straightforward tasks. Besides, Deep RL requires learning the internal states through returns implicitly \cite{matiisen2018deep}, being slow, sensitive to hyperparameters, and inefficient in practice \cite{lee2020stochastic}. A tractable alternative is to learn based on a well-predefined model (also called model-based methods) with explicit internal states. The use of explicit internal states makes the model definition interpretable and data utilization efficient \cite{chen2020end}. Typical model-based approaches integrated with internal states include MDP \cite{kochenderfer2015decision}, POMDP \cite{ bai2015intention, hausknecht2015deep, song2016intention}, and hidden Markov models (HMM) \cite{baum1972inequality}. POMDP requires encoding the complete historical information\footnote[1]{Historical information can be encoded by recalling past features \cite{mccallum1993overcoming} or inferring the distribution over possible internal states \cite{kaelbling1998planning}.} into possible \textit{internal states} and makes an appropriate decision by evaluating the current observed state value while decoding the internal states. Therefore, the implementation of well-defined internal states can improve the learning efficiency and decision performance of algorithms. 

Most existing research on internal states focuses on intention prediction of surrounding agents \cite{choudhury2007modeling, lefevre2015learning, dong2017intention, matiisen2018deep, hafner2019learning} to provide the ego vehicle in-depth knowledge of the environment. However, they mainly focused on the internal state of each vehicle independently and assumed that the ego vehicle's internal states are directly/closely related to their driving decision. All of them are subjectively defined but beyond rationality. It is also time-consuming and costly to manually specify the relevant internal states for complex dynamic environments since the flood of data and diversity in driving tasks can overwhelm human insight and analysis. 

This paper provides a probabilistically learning approach to automatically extract the internal states of the multi-vehicle \textit{interactive process} (rather than of a single vehicle's behavior), which can guide the ego vehicle to make an appropriate decision. Based on the conclusion of our previous research in \cite{wang2020social}, we here developed a probabilistic approach (i.e., HMM-GMR) to learn and reproduce the internal dynamics of merge tasks at highway on-ramps. The proposed framework combines HMM with Gaussian mixture regression (GMR) \cite{2010Learning} to leverage temporal information into dynamic processes. The GMR estimates the internal state and then predicts to verify the internal states' effectiveness further. We also compared it to GMM-GMR that does \textit{not} consider temporal information into the dynamic process.

The remainder of this paper is organized as follows. Section II reviews related works on internal states. Section III discusses the real-world data collection and processing. Section IV introduces the HMM-GMR model. Section V analyzes the results and provides further discussions. Finally, Section VI gives the conclusions.

\section{Related Works}

This section first reviews the related works of internal states, ranging from driving style and driver intention to driving maneuver. Then, their limitations and the problem to be solved are summarized.

\subsection{Internal States for Driving Style \& Driver Intention}

AVs must infer underlying states (e.g., driving styles and intents) of surrounding vehicles and their interactions to understand the environments fully \cite{bar2011probabilistic, higgs2013two, danaf2015modeling, dong2016characterizing, morton2017simultaneous, chen2019understanding, gao2020situational, ma2020reinforcement, gao2022human}. To analyze aggressive driving and predict the driver intention, researchers in \cite{danaf2015modeling} treated the driving anger as a dynamic internal state and then built a hybrid model based on HMM. The inferred internal states can encode trajectories and distinguish different driver behaviors such as passive, aggressive, tailgater, and speeder \cite{morton2017simultaneous}. Besides, the assigned internal state plays a critical role in the action selection. For example, to deeply understand the driving environment, research in \cite{chen2019understanding} applied the Latent Dirichlet Allocation (LDA) model to discover the internal states of driving habits. Some researchers \cite{ma2020reinforcement} also presented a learning-based framework to explicitly infer the internal states of surrounding vehicles (such as aggressive or conservative) using graph neural networks and demonstrated its superiority in complex scenarios such as intersections. 

The intention estimation of surrounding vehicles can help to tackle dense interactions among agents in complex traffic scenarios \cite{berndt2008continuous, kumar2013learning, bai2015intention, dong2017intention, codevilla2018end, gao2020trajectory}. For example, a multi-class support vector machine classifier combined with a Bayesian filter can predict the internal lane-change intention of surrounding drivers \cite{kumar2013learning}. In order to guarantee the safety, efficiency, and smoothness of autonomous driving, Bai \textit{et al.} proposed an intention-aware online planning approach to estimate pedestrian intentions and addressed the uncertainties in a complex environment \cite{bai2015intention}. The authors in \cite{dong2017intention} applied a probabilistic graphical model (PGM) to predict the internal intentions of surrounding vehicles in on-ramp merge scenarios. The structure of PGM allows embedding historical information and internal states into the system. Experimental results verified that the PGM-based approach can conservative personification and ensure the safety of the merging process. Considering the same observations could lead to different actions in complex scenarios (intersection or highway merge), Codevilla \textit{et al.} \cite{codevilla2018end} explicitly modelled the internal state by introducing information about the intentions and goals. In this way, the defined driver’s underlying internal state influenced the driver's subsequent actions rather than the observations.

\subsection{Internal States for Driving Maneuver/Behavior}

In a real-world setting, AVs need to understand the surroundings and know the (internal) states of their maneuvers and behaviors. Considering the underlying (or internal) states and plans, Ben-Akiva, \textit{et al.} \cite{ben2006modeling} proposed an internal choice methodology for the highway on-ramp merge tasks in congested traffic and obtained an expected performance. Besides, Choudhury \cite{choudhury2007modeling} introduced the internal plans into the decision process to address the decision-making problem in lane-change behaviors. Choudhury applied HMM to consider previous plans when making current decisions and demonstrated that ignorance of the internal states might cause an unrealistic understanding of the surrounding traffic environment. According to the internal states such as car-following, free-flow, emergency stop, the realization of a car-following maneuver consists of several actions such as acceleration, deceleration, and do-nothing \cite{koutsopoulos2012latent}. Paschalidis \textit{et al.} \cite{paschalidis2019combining} modeled the stress level of the driver as the internal state and quantified its influence on decisions. Hsiao, \textit{et al}. \cite{hsiao2019learning} trained a multi-modal policy using variational autoencoder to infer discrete internal states of different behaviors in mixed demonstrations. They verified the associated policy using the high-dimensional visual information as inputs. A multi-head network for learning internal states is also presented to predict relevant decision factors and address the limitations of high-dimensional images in data-scarce cases \cite{kargar2020efficient}. Also, Chen \textit{et al.} explained whether and how the end-to-end network policy understands and responds to the environment by proposing an interpretable Deep RL with sequential internal states \cite{chen2020interpretable}. However, this approach is a model-free model that can not explain the decision-making process as explicitly as the model-based approach.

\subsection{Summary}

The above discussion indicates that the introduction of internal states in driving (such as driving style, driver intents, driver maneuver) enables safer and more efficient algorithms for AVs. However, the internal states combined with the probabilistic learning and inference approaches would require carefully defining the internal states in advance, challenging in complex driving settings. Moreover, although the learning-based models sometimes do not need to define the number and state in advance, it requires defining the reward function accurately, which is usually a function of the (internal) states \cite{niv2019learning}. Therefore, it is necessary to develop an approach that can systematically learn, define, and infer associated internal states while preserving interpretability.

\section{Dataset and Data Processing}
\subsection{Real-World Dataset}

We utilized the data collected from the real world -- the INTERACTION dataset \cite{zhan2019interaction}, with the following considerations: 

\begin{itemize}
\item \textbf{Scenario diversity:} The data set covers great interactive driving scenarios, such as merging scenarios, intersections, and roundabouts. 
\item \textbf{Behavior diversity:} The data set collects regular and safe driving behaviors and highly interactive and complex driving behaviors, such as adversarial/irrational/near-collision maneuvers.
\item \textbf{Clear definition:} The data set contains well-defined physical information, such as vehicles' position and speed in longitudinal and lateral directions, the corresponding timestamp with the resolution of $100$ ms, agents' type (car or truck), yaw angle, as well as the length and width of vehicles.
\end{itemize}

\subsection{Data Processing} \label{subsec:data processing}
\begin{figure}[t]
\centering
\subfloat[Real scene]{\label{level1.sub.1} \includegraphics[width=0.98\linewidth]{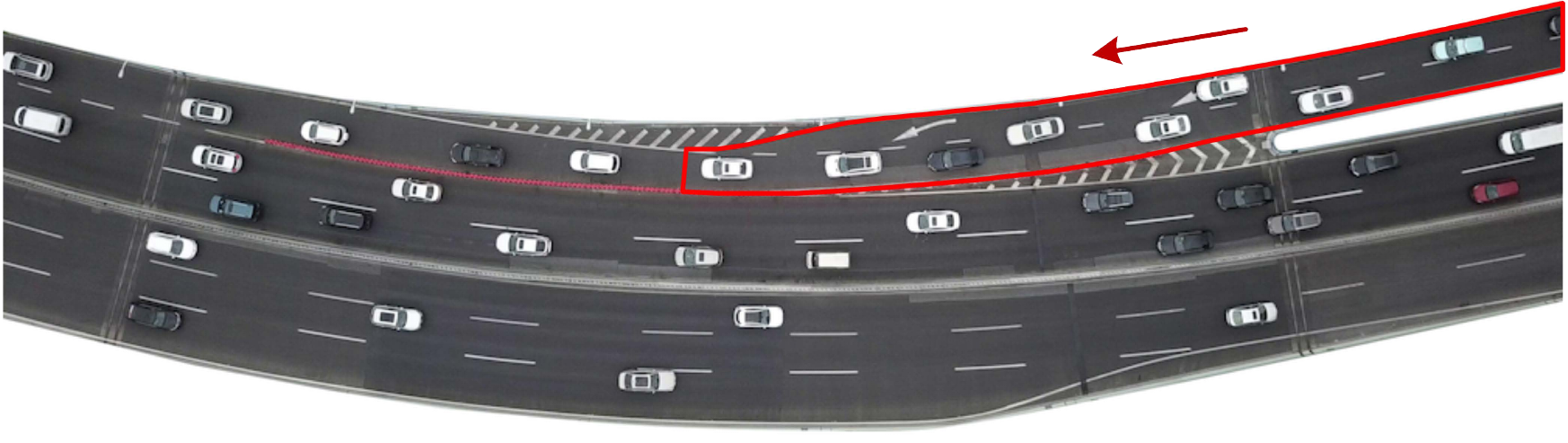}} \\
\subfloat[Data visualization]{\label{level1.sub.2} \includegraphics[width=0.98\linewidth]{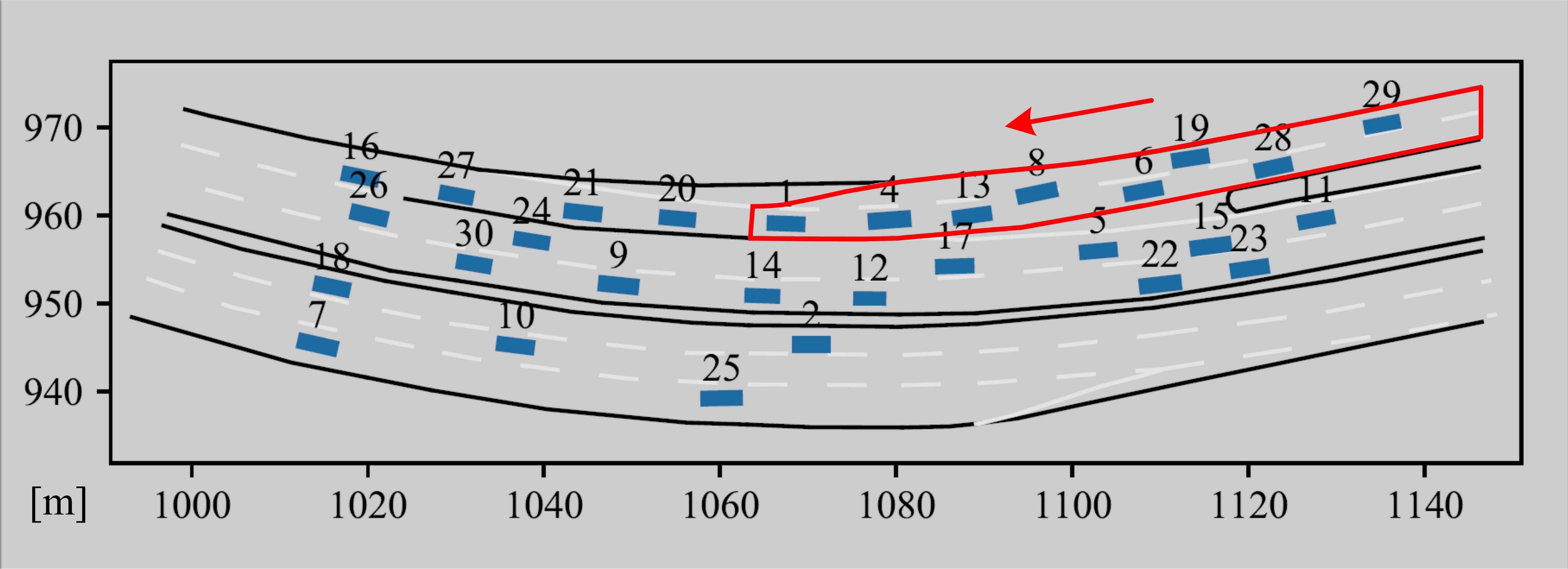}}
\caption{The Chinese highway on-ramp merge scenario in the INTERACTION dataset \cite{zhan2019interaction} and the selected local region bounded by the red line.}
\label{fig:INTERACTION dataset}
\end{figure}

The highway on-ramp merge scenarios contained in the INTERACTION dataset are from Chinese and German traffic, respectively. The video length of the Chinese (German) merge scenario is $94.62$ ($37.92$) minutes, which contains $10359$ ($574$) vehicles. As shown in Fig. \ref{fig:INTERACTION dataset}, the upper two lanes of the Chinese merge scenario is selected because they contain a longer duration and a broader variety in driving behaviors.

The data processing is based on our previous research \cite{wang2020social}. The definition of vehicles (i.e., ego vehicle, lead/lag vehicles), merge critical moments (i.e., start moment $t_s$, middle moment $t_m$, and end moment $t_e$), and social preference (rude or courteous) can refer to \cite{wang2020social}. The sequential data during the whole process with courteous merging behavior between $t_s$ and $t_e$ are extracted and the merging event amounts to 789. The extracted data are then randomly divided into a training set ($80\%$ of the dataset) and a testing set ($20\%$ of the dataset). The merge event's duration is different from each other. To make the data suitable for HMM-GMR, we screened and re-aligned the extracted data by taking a trade-off between algorithm performance and calculation capability.

\begin{figure}[t]
\centering
\includegraphics[width=\linewidth]{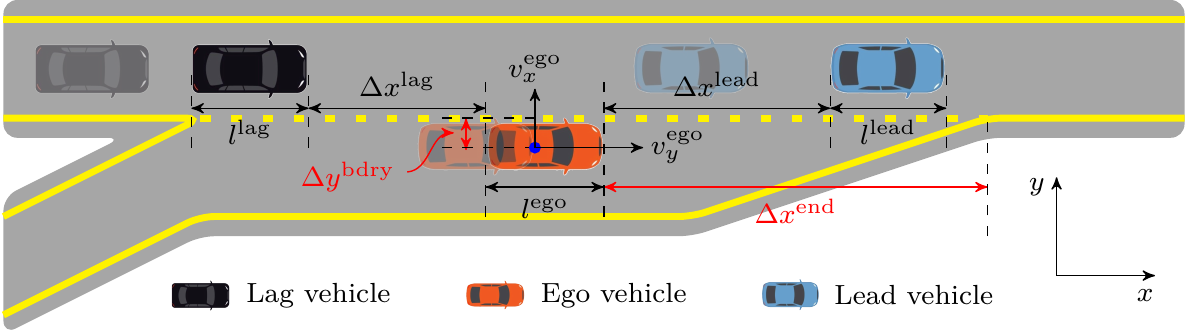}
\caption{Definition of variables for the highway on-ramp merge scenario.}
\label{fig:Diagram of a typical highway on-ramp merge scenario}
\end{figure}

The variable selection in existing works usually relies on researchers' experience and onboard sensors \cite{li2018hardware, gao2020automatic}. Our previous research \cite{wang2020social} reveals that the critical variables change over the merging stages, and redundant variables should be removed as noises. Only proper variable selection can be conducive to the inference and learning of internal states and improve decision-making performance. According to the variables defined in Fig. \ref{fig:Diagram of a typical highway on-ramp merge scenario}, different tasks require selecting different variables. For model training and internal state inference, we defined the observation at time $t$ as $\boldsymbol{x}_{t} = [\Delta v_{x}^{\mathrm{lead}}, \Delta {x}^{\mathrm{lag}}, v_x^{\mathrm{ego}}, v_y^{\mathrm{ego}}]^{\top}$. To verify the effectiveness of these learned internal states, we reconstructed some variables based on the internal state from an internal-state model and defined the inputs and outputs as
\begin{equation*}
\boldsymbol{x}_t^I =
\left[
\begin{matrix}
\Delta v_{x}^{\mathrm{lead}} \\
\Delta {x}^{\mathrm{lag}} \\
v_x^{\mathrm{ego}}
\end{matrix}
\right],
\ \ \ \ \ 
\boldsymbol{x}_t^O = v_y^{\mathrm{ego}}
\end{equation*}
The evaluation of variable selection will be given in Section \ref{subsec:evaluation of variable selection}.


\section{HMM-GMR Model} \label{sec:approach}

In this section, we developed an HMM-GMR framework to learn the internal states of the dynamic merge process from various demonstrations. We also build a probabilistic model to reproduce the sequential observations from these extracted internal states, thus verifying the model effectiveness. First, we will introduce the basis of HMM, including its framework and parameter estimation via the Baum–Welch algorithm. Then, we extended the traditional GMR to consider the spatial and sequential information contained in the HMM.

\subsection{HMM for the Merge Task}
\label{subsec:learning for HMM}
For the merge process, we assume it is subject to a Markov chain -- a mathematical model of a sequence of random variables that evolve over time in a probabilistic manner with the assumption: The value at next point in time only depends on the current state and not on what happened before. When executing complex tasks, human drivers make decisions not based on their directly-perceived signals, instead of on their unobservable internal understanding of the world. Therefore, we treated the internal modes as the discrete latent states subject to a Markov chain and the observations as the emissions of associated latent states. This operation allows formulating the merge task under a HMM framework. As a robust probabilistic method, HMM is good at dealing with spatial and temporal variabilities \cite{trabelsi2013unsupervised}. It can exhibit some degree of invariance to local warping (compression and stretching) of the time axis. A typical HMM is built on a discrete Markov model with a set of finite discrete latent states $z_t \in \mathcal{Z} = \{1,...,K\}$ and an associated observation model $p(\boldsymbol{x}_t|z_t)$. At time $t$, the observed state $\boldsymbol{x}_t$, which only depends on the current latent state $z_t$ at time $t$, is expressed as a Gaussian distribution
\begin{equation}\label{eq:state_HMM}
p(\boldsymbol{x}_t|z_t=k,\boldsymbol{\mu}_k, \boldsymbol{\Sigma}_k) \sim \mathcal{N} \left(\boldsymbol{x}_t|\boldsymbol{\mu}_k, \boldsymbol{\Sigma}_k\right)
\end{equation}
where $\boldsymbol{\mu}_k$ and $\boldsymbol{\Sigma}_k$ represent the center vector and the covariance matrix of the $k$-th Gaussian distribution, respectively. Formulating the observation model as a Gaussian distribution is intuitive with the facts: Agents do not behave directly upon their sensory data because that data is merely an indirect observation of a hidden real-world \cite{wu2020rational}, and the Gaussian distribution can be treated as a probabilistic model with latent states\cite{bishop2006pattern}. The Gaussian model parameter estimation is through the Maximum Likelihood Estimate (MLE).

Given the sequential observations $\boldsymbol{X}= \boldsymbol{x}_{1:T}$\footnote{We denote $\boldsymbol{x}_{1:T}$ as the shorthand of the sequence $\{\boldsymbol{x}_1,..., \boldsymbol{x}_T\}$.} and associated latent states $\boldsymbol{Z}= z_{1:T}$ with the Markov chain assumption, their joint probability distribution is derived by
\begin{equation}\label{eq:jpd_HMM}
p(\boldsymbol{X,Z}|\boldsymbol{\theta}) = p(z_1|\boldsymbol{\pi})\left[\prod_{t=2}^T p(z_t|z_{t-1},\boldsymbol{A})\right]\prod_{l=1}^T p(\boldsymbol{x}_l|z_l,\boldsymbol{\mu},\boldsymbol{\Sigma})
\end{equation}
where unknown model parameters $ \boldsymbol{\theta}=\{\boldsymbol{\pi}, \boldsymbol{A},\boldsymbol{\mu}, \boldsymbol{\Sigma}\} $ need to be learned. $\boldsymbol{\pi} = \{\pi_{k}\}$ is the initial probability, the entries $\pi_k$ represent the initial probability of being in state $k$. The first observation $\boldsymbol{x}_{1}$ could be assigned to one of the set of the latent states $\mathcal{Z}$ with a categorical distribution $p(z_{1}|\boldsymbol{\pi})$. $\boldsymbol{A}$ is the transition matrix, and the entries $A_{jk}=p(z_t=k|z_{t-1}=j)$ represent the probability of categorizing the current observation at time $t$ as state $k$ given the last observation at time $t-1$ being in state $j$ with $0 \leq A_{jk} \leq 1$ with $\sum_j A_{jk}=1$. Thus, $\boldsymbol{A}$ can be denoted as 
\begin{equation}\label{eq:A}
\boldsymbol{A} = 
    \begin{pmatrix}
        A_{11} & \cdots & A_{1K} \\
        \vdots & \ddots & \vdots \\
        A_{K1} & \cdots & A_{KK} \\
    \end{pmatrix}
\end{equation}

The procedure in (\ref{eq:jpd_HMM}) for the case of HMM is modified as follows. The corresponding observation $\boldsymbol{x}_1$ can be sampled based on the initial driving latent state $z_1$ with probabilities governed by $\pi_k$. The latent state of the next moment $z_2$ can be obtained according to the transition probabilities $p(z_2|z_1,\boldsymbol{A})$. Then, a sample for $\boldsymbol{x}_2$ and also $z_3$ can be drawn and so on. According to the generative procedure, our task becomes to estimate the probability of latent state sequences $z_{1:T}$ and the value of $\boldsymbol{\theta}$ that can best describe associated observation sequence $ \boldsymbol{x}_{1:T} $. The following section will detail the related algorithms.

\subsection{Parameter Learning}
For a probabilistic model estimation with latent states involved, an effective way is to conduct estimation iteratively. One typical approach is the expectation-maximization (EM) algorithm which performs the maximum likelihood estimation of HMM. It alternates between estimating the values of latent states (E-step) and optimizing the model (M-step), then repeating these two steps until convergence. As a variant of the EM algorithm, the Baum–Welch algorithm \cite{baum1970maximization, rabiner1989tutorial} can evaluate the parameters $\boldsymbol{\theta}$ of HMM efficiently.

\subsubsection{E-Step}
In the E-step, we fixed the estimated model parameter at the last iteration (denoted as $\boldsymbol{\theta}^{\mathrm{old}}$) and then calculated the marginal probability distribution for latent state of occupying state $k$ at time $t$, denoted as $\gamma_t(k) = p(z_t=k|\boldsymbol{X},\boldsymbol{\theta}^{\mathrm{old}})$ and the posterior probability of transforming from latent state $j$ at time $t-1$ to latent state $k$ at time $t$, denoted as $\xi_t(j,k) = p(z_{t-1}=j,z_t=k|\boldsymbol{X},\boldsymbol{\theta}^{\mathrm{old}})$. First, we determined the posterior distribution of the latent states $p(\boldsymbol{Z} | \boldsymbol{X}, \boldsymbol{\theta}^{\mathrm{old}})$ based on $\boldsymbol{X}$, the observation values and $\boldsymbol{\theta}^{\mathrm{old}}$, the model parameters of the EM algorithm at last iteration. Then, we evaluated the expectation of the log-likelihood for the complete data as a function of $\boldsymbol{\theta}$ 

\begin{equation}\label{eq:Q_E}
\begin{aligned}
\mathcal{Q}(\boldsymbol{\theta},\boldsymbol{\theta}^{\mathrm{old}}) =& \sum_{\boldsymbol{Z}}p(\boldsymbol{Z} | \boldsymbol{X},\boldsymbol{\theta}^{\mathrm{old}}) \ln p(\boldsymbol{X,Z}|\boldsymbol{\theta}) \\
=& \sum_{k=1}^K \gamma_1(k)\ln\pi_k 
+ \sum_{t=2}^T\sum_{j=1}^K\sum_{k=1}^K \xi_t(j,k)\ln A_{jk} \\
& + \sum_{t=1}^T\sum_{k=1}^K \gamma_t(k)\ln p(\boldsymbol{x}_t|\boldsymbol{\mu}_k,\boldsymbol{\Sigma}_k)
\end{aligned}
\end{equation}

Here, $\gamma_t(k)$ and $\xi_t(j,k)$ are evaluated via an efficient forward-backward algorithm \cite{rabiner1989tutorial}. The forward variable $\alpha_t(k)$ accounts for the joint probability of observing all the partial observation sequence $\boldsymbol{x}_{1:t}$ up to time $t$ and occupying state $k$ at time $t$ is (see Appendix-A)
\begin{equation}\label{eq:alpha_E}
\alpha_t(k)=\mathcal{N}(\boldsymbol{x}_t|\boldsymbol{\mu}_k,\boldsymbol{\Sigma}_k) \sum_{m=1}^K \alpha_{t-1}(m)A_{mk}
\end{equation}
with $\alpha_1(k)=\pi_{k}  \mathcal{N}(\boldsymbol{x}_1|\boldsymbol{\mu}_k,\boldsymbol{\Sigma}_k)$. Similarly, the backward variable $\beta_t(k)$ accounts for the conditional probability of all the future partial observation sequence $\boldsymbol{x}_{t+1:T}$ given the state $k$ at time $t$ is (see Appendix-B)
\begin{equation}\label{eq:beta_E}
\beta_t(k) = \sum_{m=1}^K A_{km}\mathcal{N}(\boldsymbol{x}_{t+1}|\boldsymbol{\mu}_m,\boldsymbol{\Sigma}_m)\beta_{t+1}(m)
\end{equation}
with $\beta_T(k)=1$. Thus, we can separately update $\gamma_t(k)$ and $\xi_t(j,k)$ to be a probability measure, respectively, via
\begin{equation}\label{eq:gamma_E_fb}
\begin{split}
\gamma_t(k) & =  \frac{\alpha_t(k)\beta_t(k)}{\sum_{m=1}^K \alpha_t(m)\beta_t(m)} \\
\xi_t(j,k) & =\frac{\alpha_{t-1}(j)A_{jk}\mathcal{N}(\boldsymbol{x}_t|\boldsymbol{\mu}_k,\boldsymbol{\Sigma}_k)\beta_t(k)}{\sum_{m=1}^K\sum_{n=1}^K \alpha_{t-1}(m)A_{mn}\mathcal{N}(\boldsymbol{x}_t|\boldsymbol{\mu}_n,\boldsymbol{\Sigma}_n)\beta_t(n)}
\end{split}
\end{equation}

\subsubsection{M-Step}
In the M-step, we updated the parameters $\boldsymbol{\theta}$  by fixing the value of $\gamma_t(k)$ and $\xi_t(j,k)$ estimated in the E-step, thus optimizing the $\mathcal{Q}$-function alternately. More specifically, each element of $\boldsymbol{\pi}$ and $\boldsymbol{A}$ are respectively maximized by
\begin{equation}\label{eq:pi_M}
\begin{split}
\pi_k & =\frac{\gamma_1(k)}{\sum_{j=1}^K \gamma_1(j)} \\
A_{jk} & =\frac{\sum_{t=2}^T \xi_t(j,k)}{\sum_{t=2}^T \sum_{n=1}^K \xi_t(j,n)}
\end{split}
\end{equation}

Besides, equation (\ref{eq:Q_E}) shows that only its final term depends on $\boldsymbol{\mu}_k$ and $\boldsymbol{\Sigma}_k$ and has exactly the same form as the data-dependent term in the corresponding function for a standard mixture distribution for independently identically distribution data. Therefore, by maximizing the function $\mathcal{Q}(\boldsymbol{\theta},\boldsymbol{\theta}^{\mathrm{old}})$ with a weighted version of the MLE of a multivariate Gaussian, we obtain the updates of $\boldsymbol{\mu}_k$  and $\boldsymbol{\Sigma}_k$ as
\begin{equation}\label{eq:mu_M}
\boldsymbol{\mu}_k=\frac{\sum_{t=1}^T \gamma_t(k) \boldsymbol{x}_t}{\sum_{t=1}^T \gamma_t(k)}
\end{equation}

\begin{equation}\label{eq:Sigma_M}
\boldsymbol{\Sigma}_{k}=\frac{\sum_{t=1}^T \gamma_t(k)(\boldsymbol{x}_t-\boldsymbol{\mu}_k)(\boldsymbol{x}_t-\boldsymbol{\mu}_k)^{\top}}{\sum_{t=1}^T \gamma_t(k)}
\end{equation}

The above E-step and M-step are performed alternatively until convergence, and the associated parameters $\boldsymbol{\theta}$ are updated according to the latest estimation. The final obtained optimal parameters $\boldsymbol{\theta}^{\ast}$ can be used to infer the \textit{internal states} of the dynamic merge process.

\subsection{Internal States in HMM-GMR} \label{subsec:GMR for HMM}
The above section introduces the HMM method to formulate the sequential observations with \textit{latent states}. However, the learn latent states are not exactly equal to the \textit{internal states} of the dynamic interaction process. The \textit{internal states} should represent the dynamic interaction process and can rebuild and reproduce the associated behavior efficiently. Therefore, we define a probabilistic model based on the internal states to produce a distribution of associated behaviors.

Inspired by the fact that the reproduction of specific movement represented with GMMs can be formalized as a regression problem \cite{ghahramani1993supervised}, we treated the above trained HMM with Gaussian-based emissions as a Gaussian mixture with certain sequential constraints. This alteration allows utilizing the Gaussian mixture regression (GMR) to retrieve associated behavior probabilistically. The retrieval performance corresponds to the representativeness of the learned \textit{internal states}. Here, based on the learned HMM parameters $\boldsymbol{\theta}=\{\boldsymbol{\pi}, \boldsymbol{A}, \boldsymbol{\mu}, \boldsymbol{\Sigma} \}$, we need to define the dynamic process of the internal states during the merging process. For a specific observation, we assume that several finite discrete potential internal states exist to be assigned, and each of them has different possibilities. Thus, the internal state, denoted by $h_k(\boldsymbol{x}_t^I)$, can be treated as a probability measure with $\sum_{k}h_k(\boldsymbol{x}_t^I) = 1$. 

As claimed above, the appropriate internal state should be able to reproduce associated behavior precisely. Therefore, we can build a GMR model with these internal states integrated to evaluate the effectiveness. Unlike other regression methods such as artificial neural networks, locally weighted regression, and locally weighted projection regression, the GMR derives the regression function from the joint probability density function of the data rather than modeling the regression directly \cite{2010Learning, shao2019bayesian}. The model training is then carried out offline, linearly related to the number of data points. The calculation of GMR is faster than other regression algorithms. Besides, GMR can handle multi-dimensional input and output variables under the same model architecture. 

For the merge task at highway on-ramps, a critical variable that can reflect the driver intent is the ego vehicle's lateral speed, $v_{y}^{\mathrm{ego}}$: A high (low) lateral speed indicates a strong (weak) intent to merge. Therefore, we treated the variable $v_{y}^{\mathrm{ego}}$ as the output of GMR and the other variables as the inputs of GMR. In what follows, the superscripts $I$ and $O$, which represent the exponents for matrices or vectors, are used to distinguish between input and output. In what follows, we use the block decomposition of data $\boldsymbol{x}$ as
\begin{equation}\label{eq:xt_GMR}
\boldsymbol{x}=
\left[ 
\begin{matrix}
\boldsymbol{x}^I \\ \boldsymbol{x}^O 
\end{matrix}
\right]
\end{equation}
where $\boldsymbol{x}^I$ and $\boldsymbol{x}^O$ represent the inputs and output defined in Section \ref{subsec:data processing}, respectively. 

%
%
%

\begin{figure}
\centering
\includegraphics[width=\linewidth]{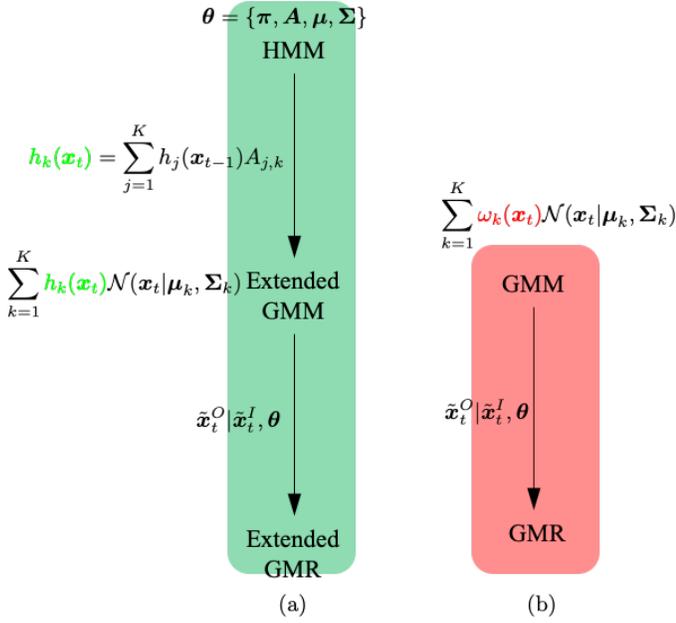}
\caption{Illustration of the frameworks of (a) HMM-GMR and (b) GMM-GMR.}
\label{fig:HMM_GMR}
\end{figure}

The representation of each observation in HMM depends on the previous choices and is jointly determined by the different components and their probabilities. Thus, HMM can be interpreted as an extended mixture model and its parameters can also be applied in GMR. More specifically, the distribution of any observation $\boldsymbol{x}$ falling in the $k$-th state of HMM can thus be expressed as a multivariate Gaussian with mean and covariance 
\begin{equation}\label{eq:musigma_HMM}
\boldsymbol{\mu}_k=\begin{bmatrix} \boldsymbol{\mu}_k^I \\ \boldsymbol{\mu}_k^O \end{bmatrix},
\boldsymbol{\Sigma}_k=\begin{bmatrix} \boldsymbol{\Sigma}_k^{II} & \boldsymbol{\Sigma}_k^{IO}\\ \boldsymbol{\Sigma}_k^{OI} & \boldsymbol{\Sigma}_k^{OO} \end{bmatrix}
\end{equation}
We decompose the mean vector and covariance matrix corresponding to the block decomposition in (\ref{eq:xt_GMR}). Equation (\ref{eq:musigma_HMM}) implies that the joint distribution of the inputs and output is a Gaussian distribution. According to \cite{murphy2012machine}, for any new input $\tilde{\boldsymbol{x}}^{I}_{t}$, the associated output $\tilde{\boldsymbol{x}}^{O}_{t}$ is also a multimodal distribution conditional on the estimated model parameters by 
\begin{equation}\label{eq:cpd_HMM}
\tilde{\boldsymbol{x}}^{O}_{t}| \tilde{\boldsymbol{x}}^{I}_{t}, \boldsymbol{\theta} \sim \sum_{k=1}^{K} h_{k}(\tilde{\boldsymbol{x}}_{t}^{I}) \mathcal{N}(\hat{\boldsymbol{\mu}}_k^{O}, \hat{\boldsymbol{\Sigma}}_k^{O|I})
\end{equation}
with the weights $h_{k}(\tilde{\boldsymbol{x}}_{t}^{I})$ and
\begin{equation}\label{eq:mu_HMM_GMR}
\begin{split}
\hat{\boldsymbol{\mu}}_k^{O}(\tilde{\boldsymbol{x}}_{t}^{I}) & = \boldsymbol{\mu}_k^O + \boldsymbol{\Sigma}_k^{OI} \left(\boldsymbol{\Sigma}_k^{II} \right)^{-1} (\tilde{\boldsymbol{x}}_t^I-\boldsymbol{\mu}_k^I) \\
\hat{\boldsymbol{\Sigma}}_k^{O|I} & = \boldsymbol{\Sigma}_k^{OO} - \boldsymbol{\Sigma}_k^{OI} \left(\boldsymbol{\Sigma}_k^{II} \right)^{-1} \boldsymbol{\Sigma}_k^{IO}
\end{split}
\end{equation}
The conditional probability distribution function of the observations is the weighted summation of different components in the mixture Gaussian at each time step $t$.

Fig. \ref{fig:HMM_GMR}(b) illustrates that the traditional development of GMR relies on a parameterized GMM, and the weights $\omega_{k}$ corresponding to each Gaussian component represent the associated influence on the input data but independent of time and sequence \cite{ghahramani1993supervised}. 
In our case, we need to first transfer the well-trained HMM with Gaussian-based emissions as an extended Gaussian mixture model. Unlike in the traditional GMM, we need to consider the influence of previous observation one step ahead on the current observation when estimating $h_{k}(\boldsymbol{x}_{t})$, as shown in Fig. \ref{fig:HMM_GMR}(a). The likelihood of current observation $\boldsymbol{x}_{t}$ belongs to component $k$ (i.e., $h_{k}(\boldsymbol{x}_{t})$) is estimated as the expectation of the likelihood of the previous observation $\boldsymbol{x}_{t-1}$ belong to all components $j=1,\dots, K$ with a transition probability $A_{j,k}$. Therefore, to make the extended GMR like HMM leveraging the spatial and sequential information, the likelihood function $h_{k}(\tilde{\boldsymbol{x}}_{t})$ is estimated recursively with the HMM representation. Thus, the weights $h_k(\tilde{\boldsymbol{x}}_t^I)$ in (\ref{eq:cpd_HMM}) are derived as
\begin{equation}\label{eq:hk_HMM_GMR}
h_k(\tilde{\boldsymbol{x}}_t^I) = \frac{\left(\sum_{m=1}^{K} h_m(\tilde{\boldsymbol{x}}_{t-1}^I) A_{mk} \right) \mathcal{N} \left(\tilde{\boldsymbol{x}}_t^I|\boldsymbol{\mu}_k^I,\boldsymbol{\Sigma}_k^{II} \right)}{\sum_{n=1}^K \left(\sum_{m=1}^{K} h_m(\tilde{\boldsymbol{x}}_{t-1}^I) A_{mn} \right) \mathcal{N} \left(\tilde{\boldsymbol{x}}_t^I|\boldsymbol{\mu}_n^I,\boldsymbol{\Sigma}_n^{II} \right)}
\end{equation}
corresponds to the probability of observing the partial sequence $\tilde{\boldsymbol{x}}_{1:t}$ and of being in state $k$ at time $t$, where $h_k(\tilde{\boldsymbol{x}}_t^I)$ is the forward variable of HMM. When $t=1$, the initial value is set by
\begin{equation*}
h_k(\tilde{\boldsymbol{x}}_1^I) = 
\frac{\pi_k \mathcal{N} \left(\tilde{\boldsymbol{x}}_1^I | \boldsymbol{\mu}_k^I, \boldsymbol{\Sigma}_k^{II} \right)}{\sum_{n=1}^K \pi_n \mathcal{N} (\tilde{\boldsymbol{x}}_1^I | \boldsymbol{\mu}_n^I, \boldsymbol{\Sigma}_n^{II})}
\end{equation*}

Equation (\ref{eq:cpd_HMM}) provides the full predictive probability density of the HMM-GMR approach and can predict the distribution of outputs given any input. Equation (\ref{eq:hk_HMM_GMR}) is a probability measure and represents the likelihood of the current observation belong to the $k$-th Gaussian component, which can be interpreted as the human's \textit{internal} beliefs to how likely the current observation falling into the given states. 

The defined internal states' reproductive capability can evaluate their correctness. According to the definition of internal states, the expectation with the probability of the well-learned internal states should be as close as possible to the actual measurement. Therefore, we provide a point prediction result by evaluating the expectations of the estimated conditional centers $\hat{\boldsymbol{\mu}}^{O}(\tilde{\boldsymbol{x}}_t^I)$
\begin{equation}\label{eq:point_mu_HMM_GMR}
\hat{\boldsymbol{\mu}}^{O}(\tilde{\boldsymbol{x}}_t^I) = \sum_{k=1}^{K} h_k(\tilde{\boldsymbol{x}}_t^I) \hat{\boldsymbol{\mu}}_k^{O}(\tilde{\boldsymbol{x}}_t^I)
\end{equation}
A small deviation to (\ref{eq:point_mu_HMM_GMR}) indicates a good performance.

\section{Result Analysis and Discussion}

This section first introduces the structure learning for HMM and then defines two evaluation metrics to assess the variable selection and the HMM-GMR performance. Afterward, the analysis of learned internal states and related potential applications are provided.

\subsection{Model Selection}

The Baum–Welch algorithm (introduced in Section \ref{sec:approach}) is a variant of the EM algorithms, which requires determining the number of components $K$ for GMM in advance. The optimal model parameter $K$ is determined by gradually increasing the number of components and selecting the optimum based on off-the-shelf criteria called Bayesian information criteria (BIC) \cite{schwarz1978estimating, watanabe2013widely}, balancing the model's likelihood and the minimum parameter number. The computation of the BIC score is given by
\begin{equation}\label{eq:BIC}
S_{\mathrm{BIC}}=-\sum_{t=1}^T \mathrm{log}(p(\boldsymbol{x}_t))+\frac{n_p}{2} \mathrm{log}(T)
\end{equation}
where the first term represents the log-likelihood and determines the fitting level of the model to the data. The second penalty factor realizes the number minimization of parameters with $n_p$ the number of parameters that can be calculated by polynomials about $K$. $\{\boldsymbol{x}_t\}$ is the set of training data point, and $T$ represents the number of data. 

\begin{figure}[t]
\centering
\includegraphics[width=\linewidth]{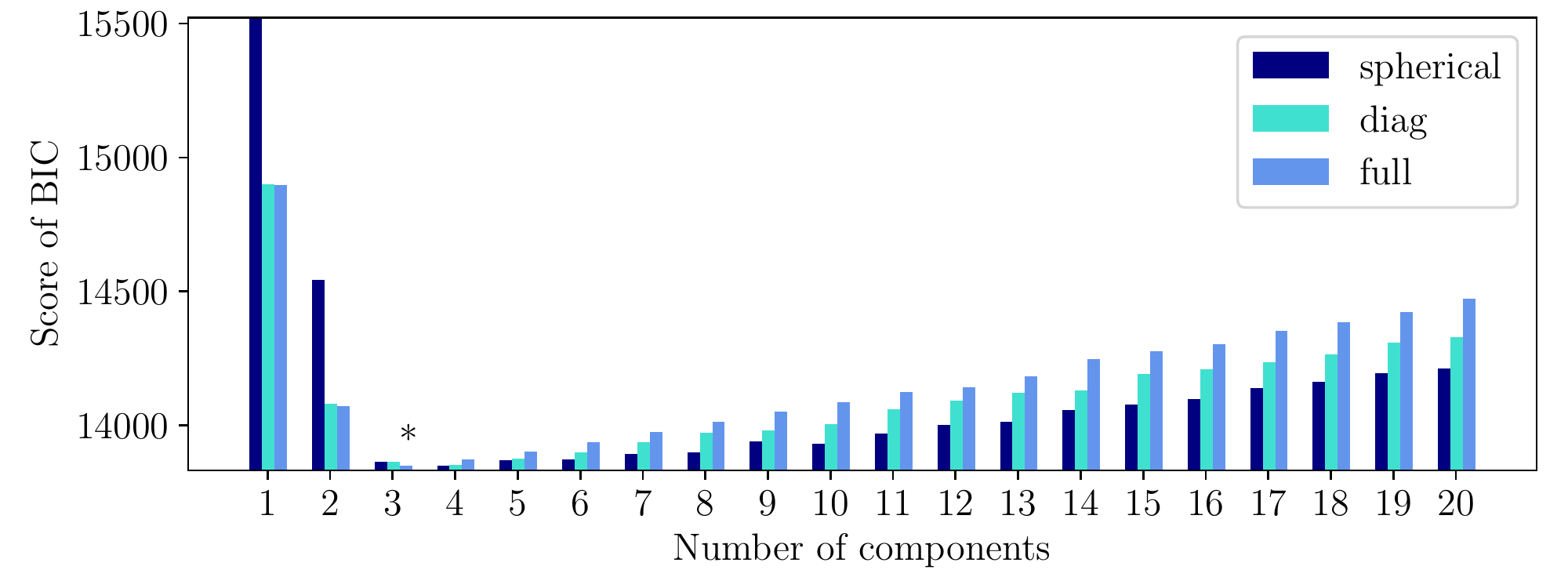}
\caption{The BIC scores of GMM with different components $K$.}
\label{fig:BIC}
\end{figure}

For the model selection, we calculate the BIC scores with different components from 1 to 20, as shown in Fig. \ref{fig:BIC}. It indicates that the BIC score first decreases and goes up with increasing $K$. This suggests that the mixture model with $K=3$ (marked with $\ast$) is the optimal selection to achieve the best performance while minimizing the parameter number. Therefore, considering the states/clusters in the framework are multivariate normal distributions with a full covariance matrix, the mixture model with $3$ Gaussian components is optimal for describing real-world driving data.

\subsection{Performance Evaluation}

The model performance is evaluated using the mean-square error (MSE) and root-mean-square error (RMSE). The MSE describes the unbiased estimation of the error variance, computed by
\begin{equation}\label{eq:MSE}
\epsilon_{\mathrm{MSE}}=\frac{1}{T}\sum_{t=1}^T (\hat{x}_t^O - x_t^{O})^2
\end{equation}
where $\hat{x}_t^O = \hat{\boldsymbol{\mu}}^{O}(\tilde{\boldsymbol{x}}_t^I)$ is the estimation of output variable at time $t$ and computed via (\ref{eq:point_mu_HMM_GMR}). $x_t^{O}$ is the real reference value collected from sensor. Therefore, the associated evaluation score of MSE is computed as \cite{mccuen2006evaluation}
\begin{equation}\label{eq:s_MSE}
S_{\mathrm{MSE}}=\frac{\epsilon_{\mathrm{MSE}} - \epsilon_{\mathrm{MSE}}^{\mathrm{ref}}}{0 - \epsilon_{\mathrm{MSE}}^{\mathrm{ref}}}
\end{equation}
with $\epsilon_{\mathrm{MSE}}^{\mathrm{ref}}=\frac{1}{T} \sum_{t=1}^T (\bar{x} - x_t^{O})^2$. Taking the MSE of $\bar{x}$ as the reference, the overall performance can be evaluated by the score of $S_{\mathrm{MSE}}$ which is positive (negative) if the predictive outputs is better (worser) than $\epsilon_{\mathrm{MSE}}^{\mathrm{ref}}$ while the absolute value of the score is proportional to the amplitude. 

In addition, we use RMSE as another evaluation metric, computed by
\begin{equation}\label{eq:RMSE}
\epsilon_{\mathrm{RMSE}} = \sqrt{\epsilon_{\mathrm{MSE}}}
\end{equation}
Thus, the mean values of two evaluation metrics ($\overline{S}_{\mathrm{MSE}}$ and $\overline{\epsilon}_{\mathrm{RMSE}}$) are used to evaluate the prediction stability and accuracy of the HMM-GMR performance. A high value of $\overline{S}_{\mathrm{MSE}}$ (or a low value of $\overline{\epsilon}_{\mathrm{RMSE}}$) indicates a good model performance.

\subsection{Evaluation of Variable Selection} \label{subsec:evaluation of variable selection}

The selection of appropriate input variables can eliminate the interference of redundant variables and maximize the performance of the HMM-GMR framework. Our previous research in \cite{wang2020social} reveals that the most critical variables of making decisions when merging into the highway are ranked as $v_y^{\mathrm{ego}}$, $v_x^{\mathrm{ego}}$, $\Delta v_{x}^{\mathrm{lead}}$, $\Delta {x}^{\mathrm{lag}}$, $\Delta v_{x}^{\mathrm{lag}}$, and $\Delta {x}^{\mathrm{lead}}$ ($TTC$ is not considered because the correlation between it and task execution is weak and unstable). $v_y^{\mathrm{ego}}$ represents the lateral control of the ego vehicle (i.e., the predictive outputs of HMM-GMR); thus, we only selected the other five variables as the model input candidates. 

With well-defined evaluation metrics, we compute the evaluation scores of models with different variable inputs. Here, we mainly consider the univariate input of the top-three significant variables (i.e., $\Delta v_{x}^{\mathrm{lead}}$, $\Delta {x}^{\mathrm{lag}}$ and $v_x^{\mathrm{ego}}$). For those variables with low significance, we only discuss the influence of different combinations of them with the optimal input on model performance. Table \ref{table:score_variable} summarizes the associated results by combining different variables. It shows that the input variables as the combination of $\{\Delta v_{x}^{\mathrm{lead}}, \Delta {x}^{\mathrm{lag}}, v_x^{\mathrm{ego}} \}$ is the best choice with the highest value of $\overline{S}_{\mathrm{MSE}}$ and the lowest value of  $\overline{\epsilon}_{\mathrm{RMSE}}$. 

Besides, we also investigated the other combinations, provided as follows: 
\begin{itemize}
\item \textbf{Combining univariate inputs:} Table \ref{table:score_variable} indicates that $\Delta v_{x}^{\mathrm{lead}}$ reaches a much higher value of $\overline{S}_{\mathrm{MSE}}$ than the other two combinations. However, $\Delta v_{x}^{\mathrm{lead}}$ obtains a close value of $\overline{\epsilon}_{\mathrm{RMSE}}$ to $\Delta {x}^{\mathrm{lag}}$, which are both far worse than $v_x^{\mathrm{ego}}$.
\item \textbf{Combining bivariate inputs:} The combination of $\Delta {x}^{\mathrm{lag}}$ and $v_x^{\mathrm{ego}}$ obtains the best performance with the highest value of $\overline{S}_{\mathrm{MSE}}$ and the lowest value of $\overline{\epsilon}_{\mathrm{RMSE}}$. $\Delta v_{x}^{\mathrm{lead}}$ is excluded in this case because the the bivariate inputs' coupling effect is different from the univariate inputs.
\end{itemize}
The evaluation scores of the univariate and bivariate inputs reveal that insufficient inputs can undermine model performance. 

To further confirm the reliability of the optimal combination of the three variables as inputs, we also analyzed the effects of the optimal combination (i.e., $\Delta v_{x}^{\mathrm{lead}}, \Delta {x}^{\mathrm{lag}}, v_x^{\mathrm{ego}}$) with other low-significant variables (i.e., $\Delta v_{x}^{\mathrm{lag}}, \Delta {x}^{\mathrm{lead}}$). The associated results in Table \ref{table:score_variable} show that the introduction of low-significant variables would undermine the model performance gradually since $\overline{S}_{\mathrm{MSE}}$ gets a reduced value, and $\overline{\epsilon}_{\mathrm{RMSE}}$ gets an increased value. Especially for the last case in Table \ref{table:score_variable}, the value of $\overline{S}_{\mathrm{MSE}}$ declines when considering the two low-significant variables mentioned above. This phenomenon supports the conclusion of variable significance analysis in \cite{wang2020social}, implying that considering low-significant variables will impair model performance. Therefore, it is necessary to filter redundant variables in the environment to improve the model performance. 

\renewcommand\arraystretch{1.5} 
\begin{table}[t]
	\centering
	\caption{Performance Evaluation of Different Variables with Same Approach (HMM-GMR ($K$-bins)) and Output Variable ($v_y^{\mathrm{ego}}$)}\label{table:score_variable}

	\begin{tabular}{ c | c c }

		\hline \hline
		Input variables & $\overline{S}_{\mathrm{MSE}}$ & $\overline{\epsilon}_{\mathrm{RMSE}}$ \\
		
		\hline
		$\Delta v_{x}^{\mathrm{lead}}$ & 0.346 & 0.608 \\
		
		\hline
		$\Delta {x}^{\mathrm{lag}}$ & -1.665 & 0.685 \\
		
		\hline
		$v_x^{\mathrm{ego}}$ & -0.261 & 0.124 \\
		
		\hline
		$\Delta v_{x}^{\mathrm{lead}}$, $\Delta {x}^{\mathrm{lag}}$ & -0.067 & 0.148 \\
		
		\hline
		$\Delta v_{x}^{\mathrm{lead}}$, $v_x^{\mathrm{ego}}$ & 0.525 & 0.075 \\
		
		\hline
		$\Delta {x}^{\mathrm{lag}}$, $v_x^{\mathrm{ego}}$ & 0.631 & 0.062 \\
		
		\hline
		$\Delta v_{x}^{\mathrm{lead}}$, $\Delta {x}^{\mathrm{lag}}$, $v_x^{\mathrm{ego}}$ & \textbf{0.686} & \textbf{0.059} \\
		
		\hline
		$\Delta v_{x}^{\mathrm{lead}}$, $\Delta {x}^{\mathrm{lag}}$, $v_x^{\mathrm{ego}}$, $\Delta v_{x}^{\mathrm{lag}}$ & 0.591 & 0.065 \\
		
		\hline
		$\Delta v_{x}^{\mathrm{lead}}$, $\Delta {x}^{\mathrm{lag}}$, $v_x^{\mathrm{ego}}$, $\Delta {x}^{\mathrm{lead}}$ & 0.458 & 0.065 \\
		
		\hline
		$\Delta v_{x}^{\mathrm{lead}}$, $\Delta {x}^{\mathrm{lag}}$, $v_x^{\mathrm{ego}}$, $\Delta v_{x}^{\mathrm{lag}}$, $\Delta {x}^{\mathrm{lead}}$ & 0.344 & 0.070 \\
		
		\hline \hline
	\end{tabular}
\end{table}

\subsection{Evaluation of Prediction Results}

The analysis of variable selection in the previous section shows that the combination of $\{\Delta v_{x}^{\mathrm{lead}}, \Delta {x}^{\mathrm{lag}}, v_x^{\mathrm{ego}} \}$ is optimal and then used to eliminate the interference of redundant variables on decision-making performance. To evaluate the proposed HMM-GMR performance, we compare it with its counterpart of GMM-GMR defined in Fig. \ref{fig:HMM_GMR}(b). Unlike the HMM-GMR, the weight coefficients $\omega$ of different Gaussian models in GMM-GMR in the iteration procedure are independent of time and sequence. Corresponding to (\ref{eq:hk_HMM_GMR}) in HMM-GMR, the activation in GMM-GMR for state $k$ at time step $t$ is defined as follows
\begin{equation}\label{eq:hk_GMM_GMR}
h_k(\boldsymbol{x}_t^{I}) = \frac{\omega_k \mathcal{N}(\boldsymbol{x}_t^I|\boldsymbol{\mu}_k^{I},\boldsymbol{\Sigma}_k^{II})}{\sum_{n=1}^{K}\omega_n \mathcal{N}(\boldsymbol{x}_t^I|\boldsymbol{\mu}_n^{I},\boldsymbol{\Sigma}_n^{II})}
\end{equation}

\renewcommand\arraystretch{1.5} 
\begin{table}[t]
	\centering
	\caption{Performance Evaluation of Different Approaches with Same Input Variables ($\Delta v_{x}^{\mathrm{lead}}$, $\Delta {x}^{\mathrm{lag}}$, $v_x^{\mathrm{ego}}$) and Output Variable $v_y^{\mathrm{ego}}$}\label{table:score_approach}

	\begin{tabular}{ c | c c }

		\hline \hline
		Approach (Initialization method) & $\overline{S}_{\mathrm{MSE}}$ & $\overline{\epsilon}_{\mathrm{RMSE}}$ \\
		
		\hline
		HMM-GMR ($K$-bins) & \textbf{0.686} & \textbf{0.059} \\
		
		\hline
		HMM-GMR ($K$-means) & 0.604 & 0.061 \\
		
		\hline
		GMM-GMR ($K$-bins) & 0.485 & 0.065 \\
		
		\hline
		GMM-GMR ($K$-means) & 0.329 & 0.066 \\
		
		\hline \hline
	\end{tabular}
\end{table}

\begin{figure}[t]
\centering
\subfloat[GMM-GMR]{\label{level2.sub.1}
\includegraphics[width=0.98\linewidth]{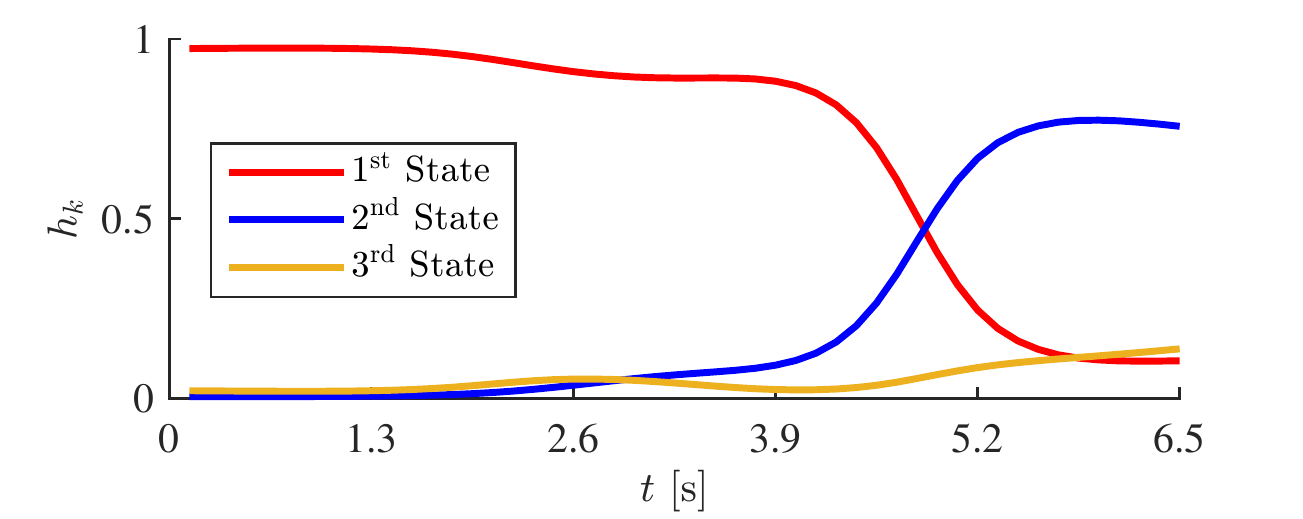}} \\
\subfloat[HMM-GMR]{\label{level2.sub.2}
\includegraphics[width=0.98\linewidth]{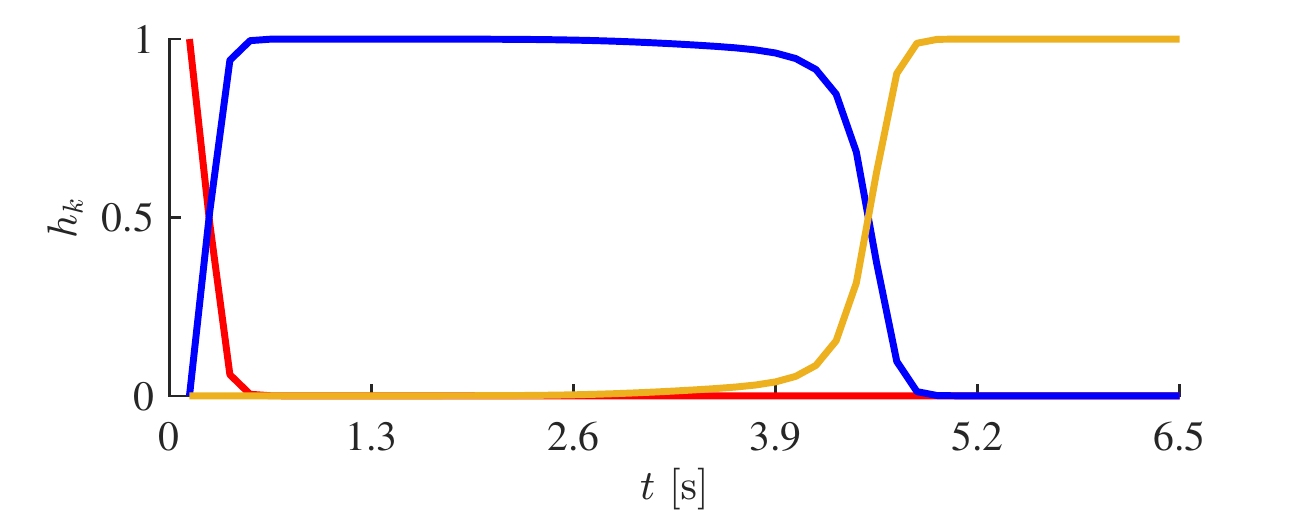}} \\
\caption{The activation weights $h_k$ (internal state) derived by GMM-GMR and HMM-GMR for one case.}
\label{fig:h_k}
\end{figure}

The model parameters should be initialized by proper initialization to avoid being trapped in poor local minima. In the training and testing processes, we introduced two different initialization methods: 
\begin{itemize}
\item $K$-means: initialize the model parameters by using $K$-means clustering algorithm; and
\item $K$-bins: initialize the model parameters by clustering an ordered dataset into equal bins.
\end{itemize} 

Table \ref{table:score_approach} displays the evaluation results and indicates that HMM-GMR outperforms GMM-GMR, reaching a higher value of $\overline{S}_{\mathrm{MSE}}$ and a lower value of $\overline{\epsilon}_{\mathrm{RMSE}}$ than GMM-GMR. Moreover, both HMM-GMR and GMM-GMR with initialization of $K$-bins always obtain a better performance than using $K$-means. Besides, Fig. \ref{fig:h_k} displays that the update of internal states based on GMM-GMR is more oscillating than HMM-GMR because GMM-GMR does not leverage the influence of time and sequence in the learning and testing phases. As a result, it leads to a large prediction error, making the internal states obtained by the activation function of GMM-GMR is not as stable as HMM-GMR. By considering the factors mentioned above comprehensively, we can conclude that the HMM-GMR framework initializing with $K$-bins obtains the best performance. This evidence proves that the internal state obtained via HMM-GMR is close to the actual situation. 

\begin{figure}[t]
\centering
\subfloat[]{\label{level3.sub.1}
\includegraphics[width=0.98\linewidth]{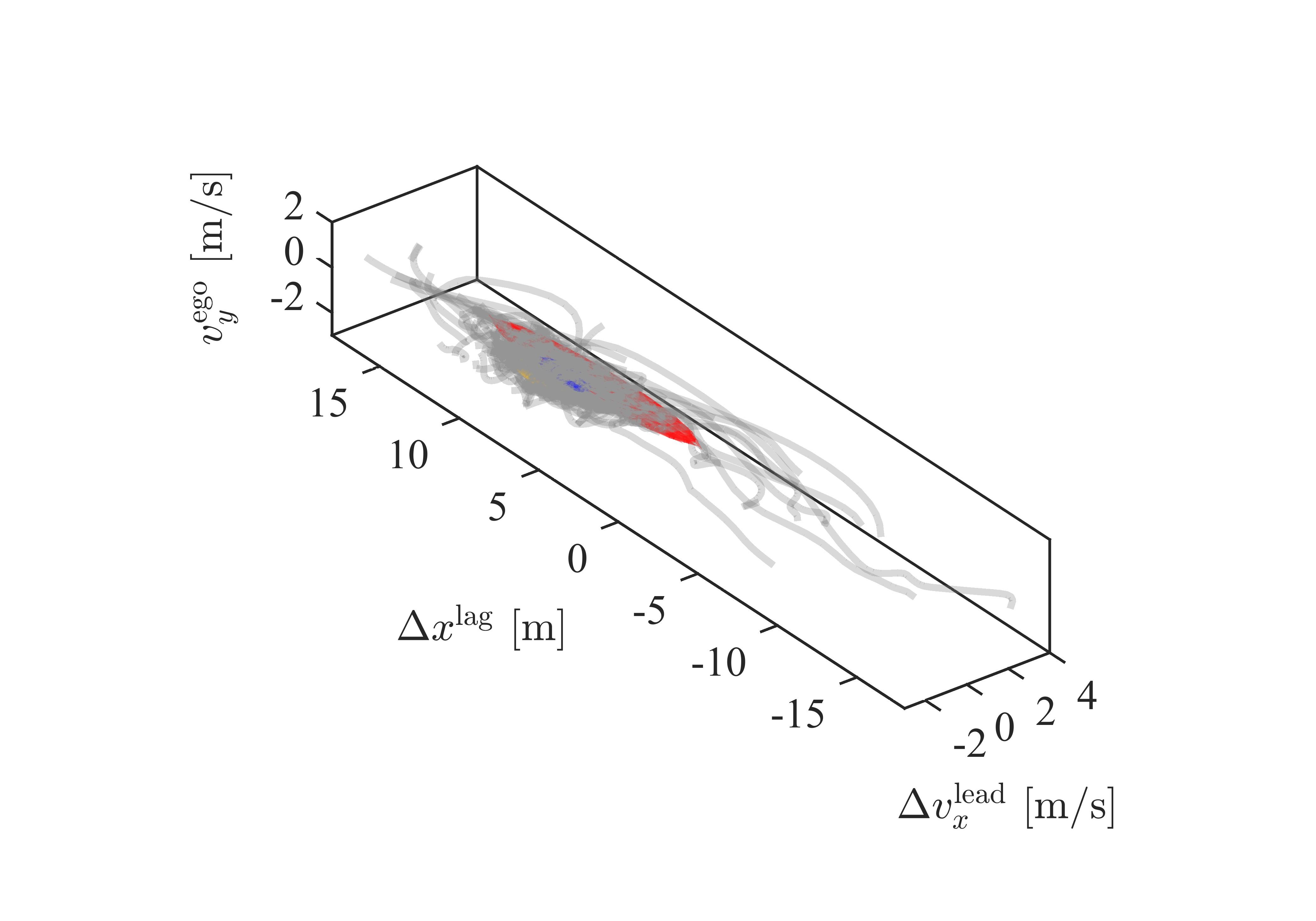}} \\
\subfloat[]{\label{level3.sub.2} \includegraphics[width=0.98\linewidth]{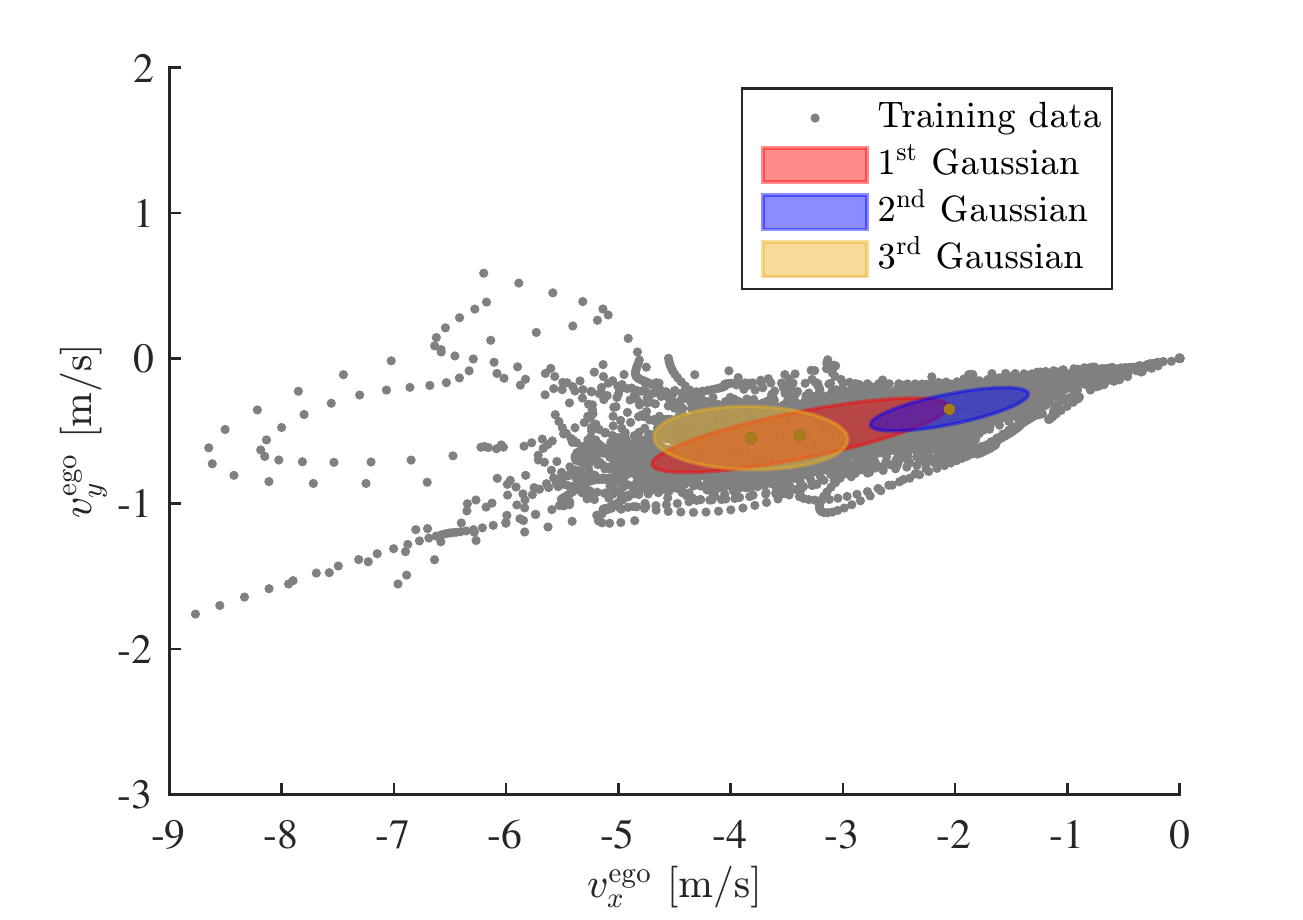}}
\caption{Training results of HMM-GMR.} 
\label{fig:Training results}
\end{figure}

\begin{figure}[t]
\centering
\subfloat[]{\label{level4.sub.1}
\includegraphics[width=0.98\linewidth]{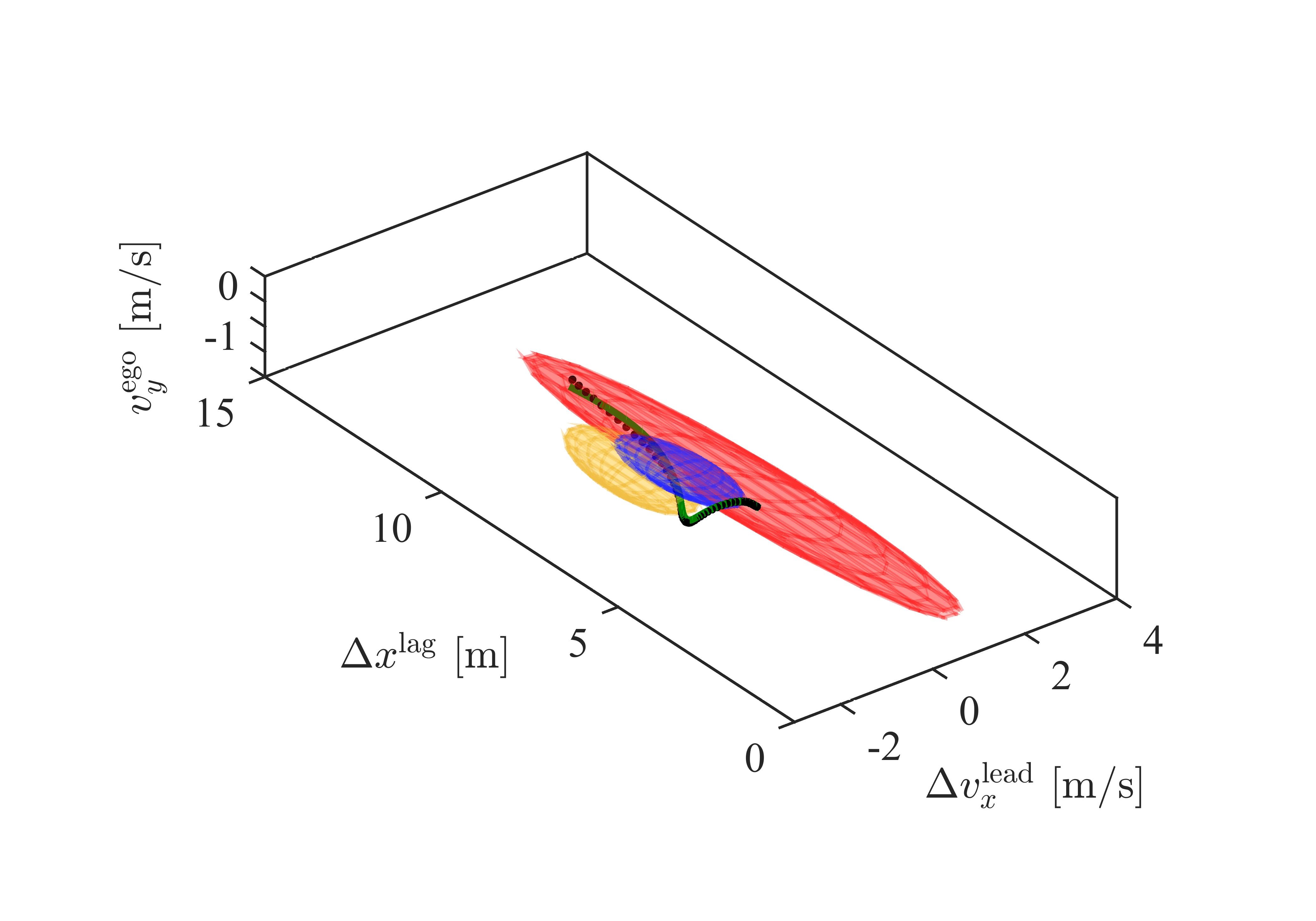}} \\
\subfloat[]{\label{level4.sub.2} \includegraphics[width=0.98\linewidth]{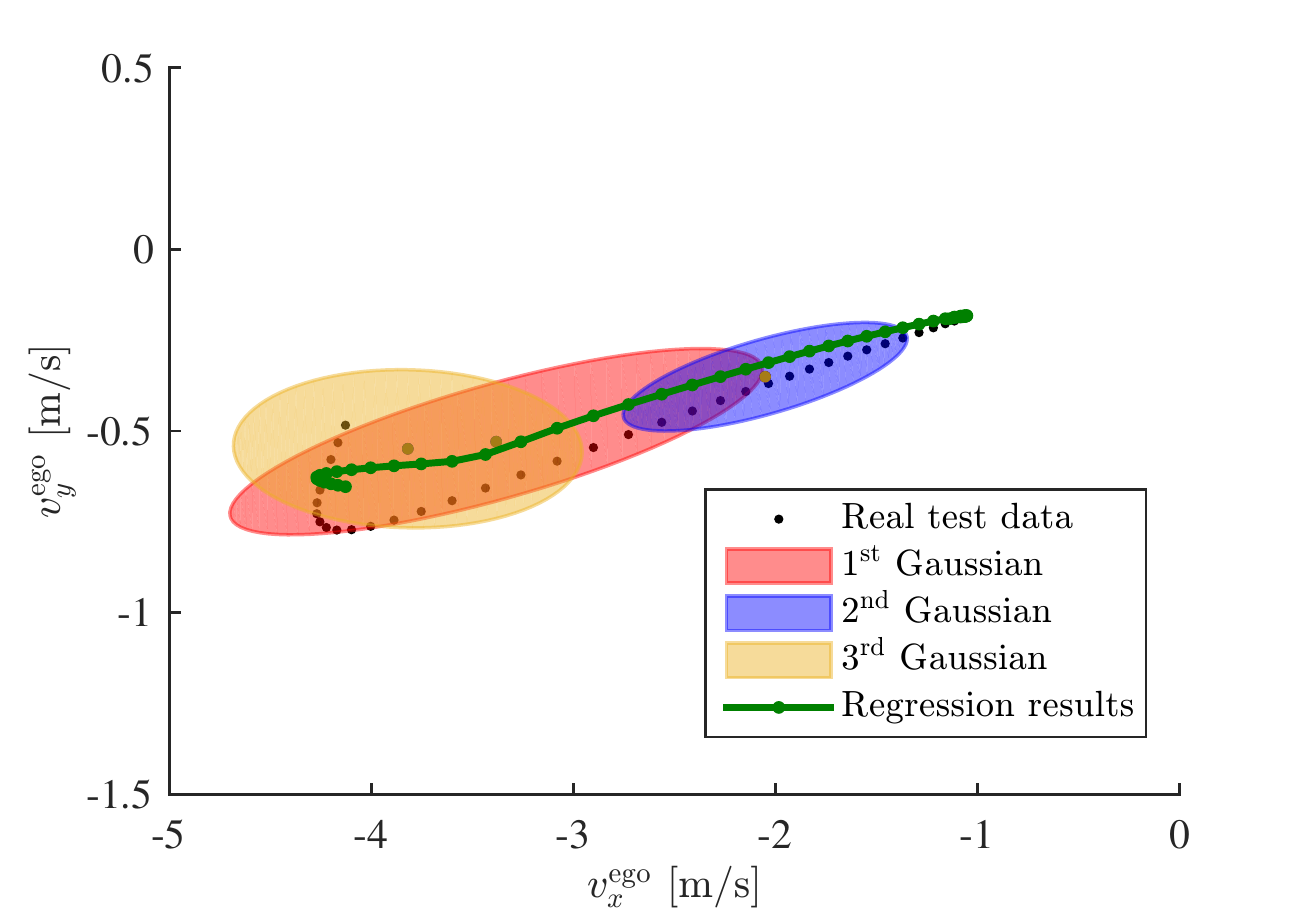}}
\caption{Example of the testing result for a randomly selected test case.} 
\label{fig:Test results}
\end{figure}

The above analysis allows treating $\{\Delta v_{x}^{\mathrm{lead}}, \Delta {x}^{\mathrm{lag}}, v_x^{\mathrm{ego}} \}$ as the inputs of HMM-GMR with $K$-bins initialization. Figs. \ref{fig:Training results} and \ref{fig:Test results} display the training (based on all the training cases) and testing (one randomly selected test case) results, respectively. Each figure shows the results from two views: two-dimensional view (bottom) and three-dimensional view (top). The two-dimensional view is a plane diagram of the relationship between the input variable $v_x^{\mathrm{ego}}$ and the output variable $v_y^{\mathrm{ego}}$. The training results (as shown in Fig. \ref{fig:Training results}) display the relationships between the three Gaussian components and all the training data, while the testing results in Fig. \ref{fig:Test results} indicate that the HMM-GMR model can obtain a good prediction performance. 

Figs. \ref{fig:Training results} and \ref{fig:Test results} display that the red Gaussian component (i.e., the first internal state) covers the most wider range over the three independent variables, while the blue Gaussian component (i.e., the second internal state) obtains the narrowest one. The randomness of the relative relationship between the ego vehicle and the surrounding agents is strong, while the relative relationship is more regular and concentrated in the second internal state. Besides, with the increase of $v_x^{\mathrm{ego}}$, the regularity of training data gradually weakens; that is, the test error increases with the increase of $v_x^{\mathrm{ego}}$. The reasons for this phenomenon are analyzed in Section \ref{subsubsec:limitations}.

\subsection{Interpretability of Internal State}

\begin{figure}[t]
\centering
\includegraphics[width=\linewidth]{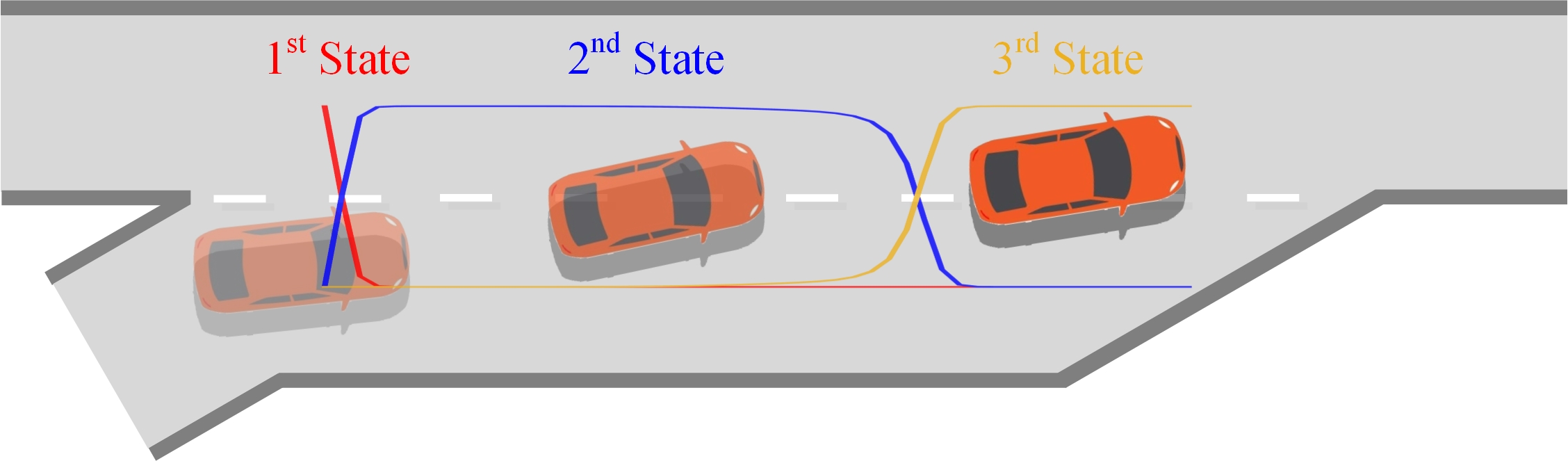}
\caption{An example of the internal state changes during the merging process.}
\label{fig:state transition}
\end{figure}

This section will interpret the learned internal states of the merging behavior at highway on-ramps semantically. According to the update of the activation coefficient in Fig. \ref{level2.sub.2}, Fig. \ref{fig:state transition} displays how the internal states correspond to the merging procedure over time.

To interpret the interaction behavior during the merging task with the three learned internal states, we listed the range of each internal state for each input variable in Table \ref{table:range}. All collected vehicle speed is non-positive because all vehicles in the dataset drive toward the left direction, as shown in Fig. \ref{fig:INTERACTION dataset}. Table \ref{table:range} indicates that the ego vehicle's absolute speed first decreases and then gradually increases. However, the speed difference between the lead vehicle and the ego vehicle decreases and finally remains non-positive from the second state. From the second state, the ego vehicle moves slower than the lead vehicle. Although the ego vehicle gradually accelerates, it always moves slower than the lead vehicle to keep a safe distance from the lead vehicle. Besides, the distance between the lag vehicles and the ego vehicles first decreases and then increases, indicating that after the second state, the rear vehicle would actively increase the safety distance to the ego vehicle. The dynamic interactions reflected by the corresponding relationship between the internal states and the selected variables are consistent with the highway on-ramp merge behavior in reality. Therefore, the three learned internal states can fully and concretely explain the interactive merge behavior.

\renewcommand\arraystretch{1.5} 
\begin{table}[t]
	\centering
	\caption{The Ranges of Input Variables in Different Internal States} \label{table:range}
	
	\begin{tabular}{p{0.1\columnwidth} p{0.1\columnwidth} p{0.16\columnwidth} p{0.16\columnwidth} p{0.16\columnwidth}}

		\hline \hline
		
		& & $1^{\mathrm{st}}$ State & $2^{\mathrm{nd}}$ State & $3^{\mathrm{rd}}$ State \\
		\hline
		
		$\Delta v_{x}^{\mathrm{lead}}$ & [m/s] & [-0.7, 1.2] & [-1.1, 0.0] & [-1.7, -0.2] \\
		\hline
		
		$v_x^{\mathrm{ego}}$ & [m/s] & [-4.7, -2.0] & [-2.8, -1.4] & [-4.7, -2.9] \\
		\hline
		
		$\Delta {x}^{\mathrm{lag}}$ & [m] & [-0.4, 11.6] & [4.8, 8.2] & [5.4, 9.1] \\

		\hline	\hline
	\end{tabular}
\end{table}

\subsection{Further Discussions}

\subsubsection{Potential Applications}

Model-based RL and POMDP receive increasing attention in recent years \cite{bai2015intention, hausknecht2015deep, song2016intention} in light of their interpretability and generalizability \cite{chen2020end}. POMDP usually treats the unobservable environmental uncertainty as internal states or considers the complete historical information encoded by recalling past features and inferring to determine the distribution over possible internal states \cite{kaelbling1998planning}. Although the belief state's rationality in POMDP has found some evidence in recent experimental studies \cite{babayan2018belief}, the update of this belief state requires the state transition and observation function. The HMM-GMR framework developed in this paper can provide the basic parameter update procedures for model-based approaches to improve learning efficiency and decision performance.

\subsubsection{Limitations}\label{subsubsec:limitations}

Most highway on-ramp merging scenarios in the INTERACTION dataset are in a congested, highly interactive condition. The developed HMM-GMR framework obtains an expected prediction performance and infers the internal states during the decision process. However, the model trained with this kind of data may not be suitable for the free-flow traffic conditions, which could be future work. 

\section{Conclusion}

This paper developed a probabilistic learning approach, HMM-GMR, to extract the interpretable internal states for the dynamic interaction procedure of merging at highway on-ramps. Related parameter estimation algorithms for the HMM-GMR model are provided. Experiments on the real-world data demonstrate its efficiency and reveal that the interaction procedure for merge behavior at highway on-ramps can be semantically described via three internal states. We also evaluated the HMM-GMR model with different variables as inputs. We demonstrated that the optimal model inputs are $\{\Delta v_{x}^{\mathrm{lead}}, \Delta {x}^{\mathrm{lag}}, v_x^{\mathrm{ego}} \}$ to make an appropriate decision. Moreover, the developed HMM-GMR model, to some extent, provides reliable and experimental support to the conclusions in our previous work \cite{wang2020social}.

\section*{Appendix}

\subsection{Derivation of Forward Variable $\alpha_t(k)$} \label{subsec:forward}

For the forward variable $\alpha_{t}(k)$, its estimation is based on the old parameter $\boldsymbol{\theta}^{\mathrm{old}}$, i.e., $\alpha_t(k)=p(\boldsymbol{x}_{1:t},z_t=k|{\boldsymbol{\theta}^{\mathrm{old}}})$. To simplify the proof representation, we omitted the parameter ${\boldsymbol{\theta}^{\mathrm{old}}}$ and default $p(\boldsymbol{x}_{1:t},z_t=k|{\boldsymbol{\theta}^{\mathrm{old}}}) = p(\boldsymbol{x}_{1:t},z_t=k)$. 

\begin{align*}
\begin{split}
\alpha_t(k)=&p(\boldsymbol{x}_{1:t},z_t=k)\\
=&p(\boldsymbol{x}_t|z_t=k) p(\boldsymbol{x}_{1:t-1} | z_{t}=k)p(z_t=k)\\
=&p(\boldsymbol{x}_t|z_t=k)p(\boldsymbol{x}_{1:t-1}, z_{t}=k)\\
=&p(\boldsymbol{x}_t|z_t=k) \sum_{m=1}^K p(\boldsymbol{x}_{1:t-1}, z_{t-1}=m, z_{t}=k)\\
=&p(\boldsymbol{x}_t|z_t=k) \sum_{m=1}^K [ p(\boldsymbol{x}_{1:t-1}, z_{t}=k | z_{t-1}=m) \cdot \\
 &\qquad \qquad \qquad \quad \; \, p(z_{t-1}=m) ]\\
=&p(\boldsymbol{x}_t|z_t=k) \sum_{m=1}^K [ p(\boldsymbol{x}_{1:t-1},z_{t-1}=m) \cdot \\
 &\qquad \qquad \qquad \quad \; \, p(z_{t}=k|z_{t-1}=m) ]\\
=&p(\boldsymbol{x}_t|z_t=k,\boldsymbol{\mu}_k,\boldsymbol{\Sigma}_k) \sum_{m=1}^K \alpha_{t-1}(m)A_{mk} \\
=&\mathcal{N}(\boldsymbol{x}_t|\boldsymbol{\mu}_k,\boldsymbol{\Sigma}_k) \sum_{m=1}^K \alpha_{t-1}(m)A_{mk}
\end{split}
\end{align*}

\subsection{Derivation of Backward Variable $\beta_t(k)$} \label{subsec:backward}

The estimation of backward variable $\beta_{t}(k)$ is based on the old parameter $\boldsymbol{\theta}^{\mathrm{old}}$, i.e., $\beta_t(k)=p(\boldsymbol{x}_{t+1:T}|z_t=k, {\boldsymbol{\theta}^{\mathrm{old}}})$. To simplify the proof representation, we omitted the parameter ${\boldsymbol{\theta}^{\mathrm{old}}}$ and default $p(\boldsymbol{x}_{t+1:T} | z_t=k, {\boldsymbol{\theta}^{\mathrm{old}}}) = p(\boldsymbol{x}_{t+1:T} | z_t=k)$.

\begin{align*}
\beta_t(k)=&p(\boldsymbol{x}_{t+1:T}|z_t=k)\\
=& \sum_{m=1}^K p(\boldsymbol{x}_{t+1:T},z_{t+1}=m|z_t=k)\\
=&\sum_{m=1}^K [ p(z_{t+1}=m | z_t=k) \cdot \\
 &\qquad \, p(\boldsymbol{x}_{t+1:T} | z_t=k,z_{t+1}=m) ] \\
=&\sum_{m=1}^K [ p(z_{t+1}=m | z_t=k) p(\boldsymbol{x}_{t+1:T} | z_{t+1}=m) ] \\
=&\sum_{m=1}^K [ p(z_{t+1}=m | z_t=k) p(\boldsymbol{x}_{t+1} | z_{t+1}=m) \cdot \\
 &\qquad \, p(\boldsymbol{x}_{t+2:T} | z_{t+1}=m) ] \\
=&\sum_{m=1}^K A_{km}p(\boldsymbol{x}_{t+1}|z_{t+1}=m,\boldsymbol{\mu}_m,\boldsymbol{\Sigma}_m)\beta_{t+1}(m)\\
=&\sum_{m=1}^K A_{km}\mathcal{N}(\boldsymbol{x}_{t+1}|\boldsymbol{\mu}_m,\boldsymbol{\Sigma}_m)\beta_{t+1}(m)
\end{align*}





\ifCLASSOPTIONcaptionsoff
  \newpage
\fi




\bibliographystyle{IEEEtran.bst}
\bibliography{reference}

%

%

\begin{IEEEbiography}[{\includegraphics[width=1in,height=1.25in,clip,keepaspectratio]{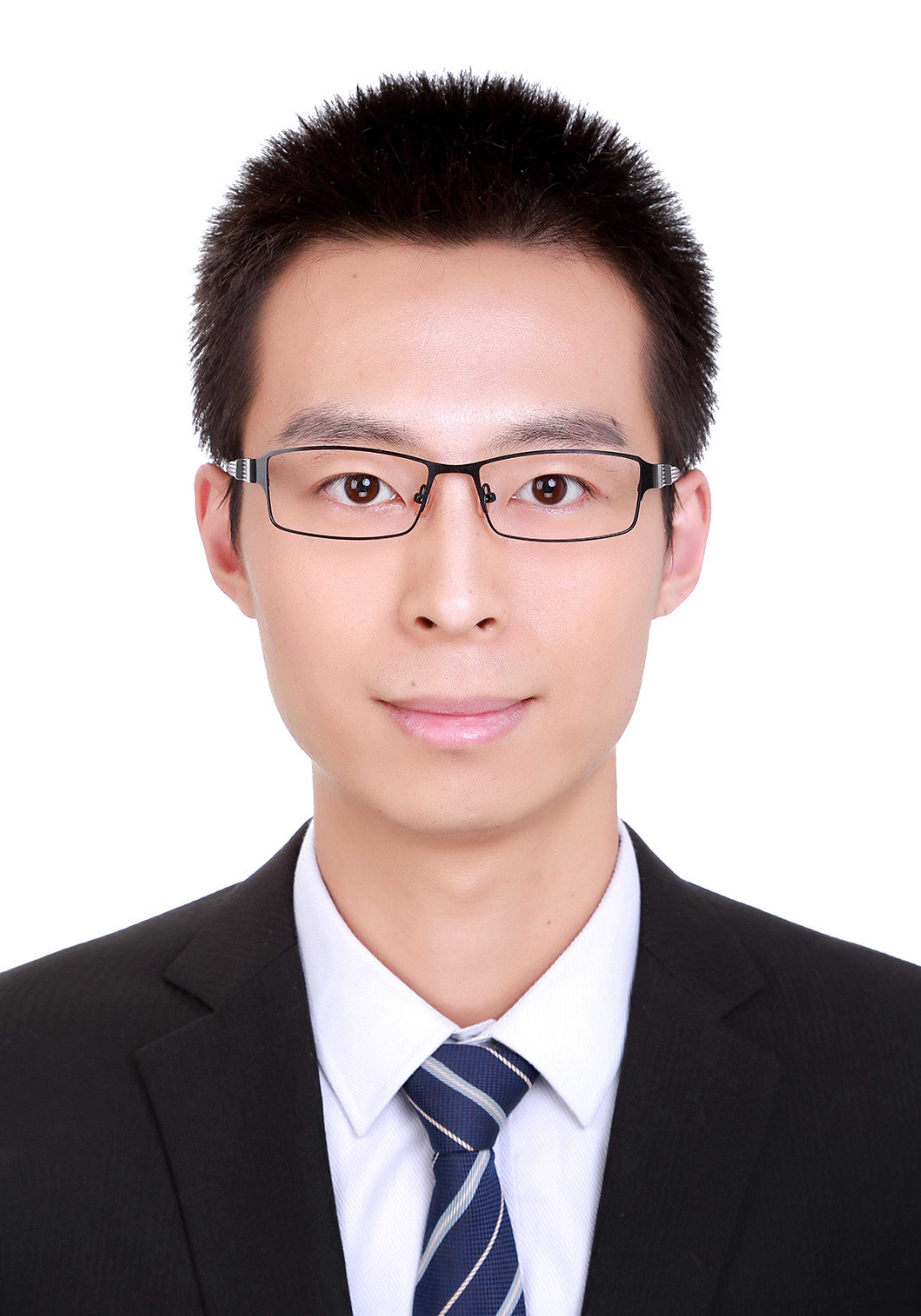}}]{Huanjie Wang}
received the M.A. degree from the School of Mechanical Engineering, Beijing Institute of Technology (BIT), China, in 2016, where he is currently pursuing the Ph.D. degree in mechanical engineering. He was also a Research Scholar with the University of California at Berkeley (UCB) from 2018 to 2020. His research interests include automated vehicle, situational awareness, decision-making, and machine learning. 
\end{IEEEbiography}

\begin{IEEEbiography}[{\includegraphics[width=1in,height=1.25in,clip,keepaspectratio]{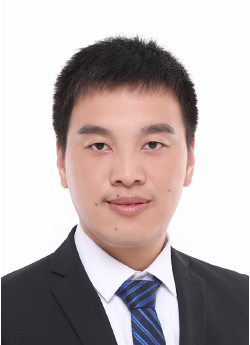}}]{Wenshuo Wang} (SM'15-M'18)
received the Ph.D. degree in mechanical engineering from the Beijing Institute of Technology, Beijing, China, in 2018. He is currently working as a Postdoctoral Researcher with California Partners for Advanced Transportation Technology (California PATH), UC Berkeley. He was a Postdoctoral Research Associate with the Carnegie Mellon University, Pittsburgh, PA, USA, from 2018 to 2019. He was also a Research Scholar with the University of California at Berkeley from 2015 to 2017 and with the University of Michigan, Ann Arbor, from 2017 to 2018. His research interests include Bayesian nonparametric learning, human driver model, human–vehicle interaction, ADAS, and autonomous vehicles.
\end{IEEEbiography}

\begin{IEEEbiography}[{\includegraphics[width=1in,height=1.25in,clip,keepaspectratio]{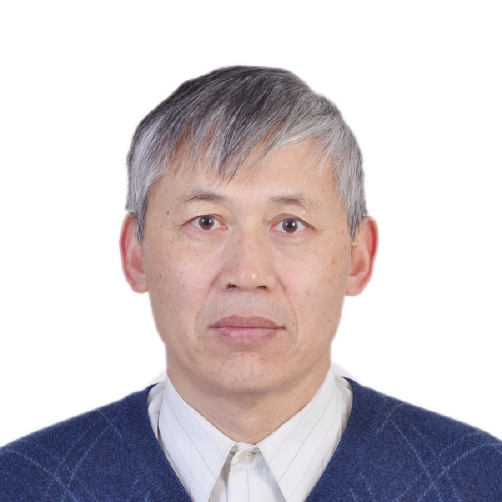}}]{Shihua Yuan}
received the B.S., M.S., and Ph.D. degrees in vehicle engineering from the Beijing Institute of Technology, Beijing, China, in 1982, 1985, and 2000, respectively. From 1992 to 1997, he was an Associate Professor with the Beijing Institute of Technology, where he has been a Professor with the School of Mechanical Engineering, since 1997. He is the author of more than 100 research articles. His research interests include vehicle dynamics, vehicle braking energy recovery, vehicle continuous transmission and its control technology, and unmanned ground vehicle.
\end{IEEEbiography}

\begin{IEEEbiography}[{\includegraphics[width=1in,height=1.25in,clip,keepaspectratio]{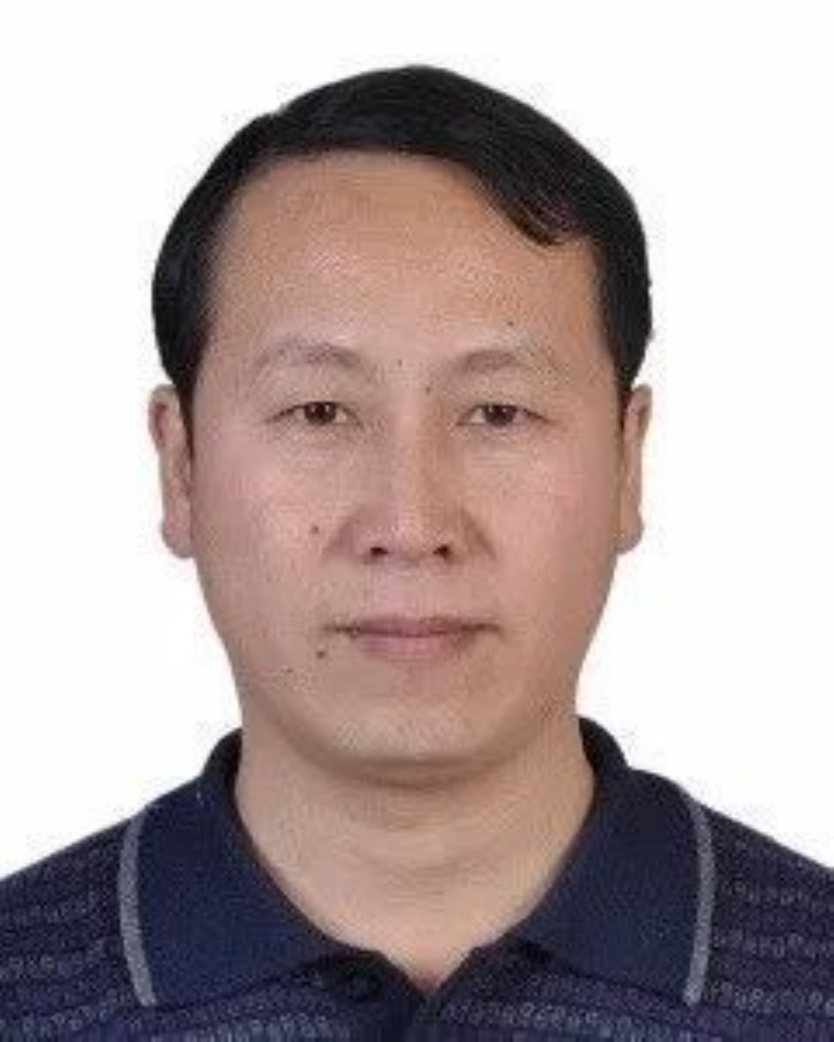}}]{Xueyuan Li}
received the B.S., M.S., and Ph.D. degrees in vehicle engineering from the Beijing Institute of Technology, Beijing, China, in 1999, 2002, and 2010, respectively. He was the Director of the National Key Laboratory of Vehicular Transmission, from 2008 to 2014. He is currently the Vice Director of the Department of Vehicle Engineering, Beijing Institute of Technology. Since 2002, he has been an Associate Professor with the School of Mechanical Engineering, Beijing Institute of Technology. His research interests include vehicle transmission theory and technology, unmanned vehicle theory and technology, and machine learning.
\end{IEEEbiography}







\end{document}